\documentclass{article} 
\usepackage{iclr2021_conference,times}

\usepackage{amsmath,amsfonts,bm}

\def\eqref#1{equation~\ref{#1}}

\def\1{\bm{1}}

\DeclareMathAlphabet{\mathsfit}{\encodingdefault}{\sfdefault}{m}{sl}
\SetMathAlphabet{\mathsfit}{bold}{\encodingdefault}{\sfdefault}{bx}{n}

\usepackage{natbib}
\usepackage{hyperref}
\usepackage{url}
\usepackage{amssymb}
\usepackage{subfigure}
\usepackage{epsfig}
\usepackage{graphicx}
\usepackage{float}
\usepackage{booktabs}

\usepackage{amsfonts}       
\usepackage{nicefrac}       
\usepackage{microtype}      

\usepackage{microtype}
\usepackage{graphicx}
\usepackage{subfigure}
\usepackage{amsmath}
\usepackage{booktabs} 
\usepackage{bm}
\usepackage{multirow}
\usepackage[misc]{ifsym} 
\iclrfinalcopy

\title{Bigeminal Priors Variational auto-encoder}

\author{
  Xuming Ran $^{1}$ ,
  Mingkun Xu$^{2} $ ,
  Qi Xu$^{3,\ast}$,
  Huihui Zhou$^{4}$,
  Quanying Liu $^{1,\ast}$}

\begin{document}
\maketitle
\vspace{-0.7cm}
\begin{centering}
$^1$ Department of Biomedical Engineering, Southern University of Science and Technology, Shenzhen 518055, P.R. China
\\
$^2$ Department of Precision Instrument, Tsinghua  University,  Beijing 100084, P. R. China \\
$^3$ College of Computer Science and Technology, Zhejiang University, Hangzhou 310027, P.R. China

$^4$ Peng Cheng Laboratory, Shenzhen 518055, P.R. China

\vspace{0.1cm}
$^\ast$ Corresponding author \texttt{liuqy@sustech.edu.cn to Q.L.; xuqi123@zju.edu.cn to Q.X.}
\end{centering}

\newcommand{\fix}{\marginpar{FIX}}
\newcommand{\new}{\marginpar{NEW}}


\begin{abstract}
Variational auto-encoders (VAEs) are an influential and generally-used class of likelihood-based generative models in unsupervised learning. The likelihood-based generative models have been reported to be highly robust to the out-of-distribution (OOD) inputs and can be a detector by assuming that the model assigns higher likelihoods to the samples from the in-distribution (ID) dataset than an OOD dataset. However, recent works reported a phenomenon that VAE recognizes some OOD samples as ID by assigning a higher likelihood to the OOD inputs compared to the one from ID. 
In this work, we introduce a new model, namely \textit{Bigeminal Priors Variational auto-encoder (BPVAE)}, to address this phenomenon. The BPVAE aims to enhance the robustness of the VAEs by combing the power of VAE with the two independent priors that belong to the training dataset and simple dataset, which complexity is lower than the training dataset, respectively. BPVAE learns two datasets’ features, assigning a higher likelihood for the training dataset than the simple dataset. In this way, we can use BPVAE’s density estimate for detecting the OOD samples. Quantitative experimental results suggest that our model has better generalization capability and stronger robustness than the standard VAEs, proving the effectiveness of the proposed approach of hybrid learning by collaborative priors. Overall, this work paves a new avenue to potentially overcome the OOD problem via multiple latent priors modeling.
\end{abstract}

\section{Introduction}

\begin{figure}[H]
\centering
\subfigure[Trained on CIFAR10]{\includegraphics[width=0.45\textwidth]{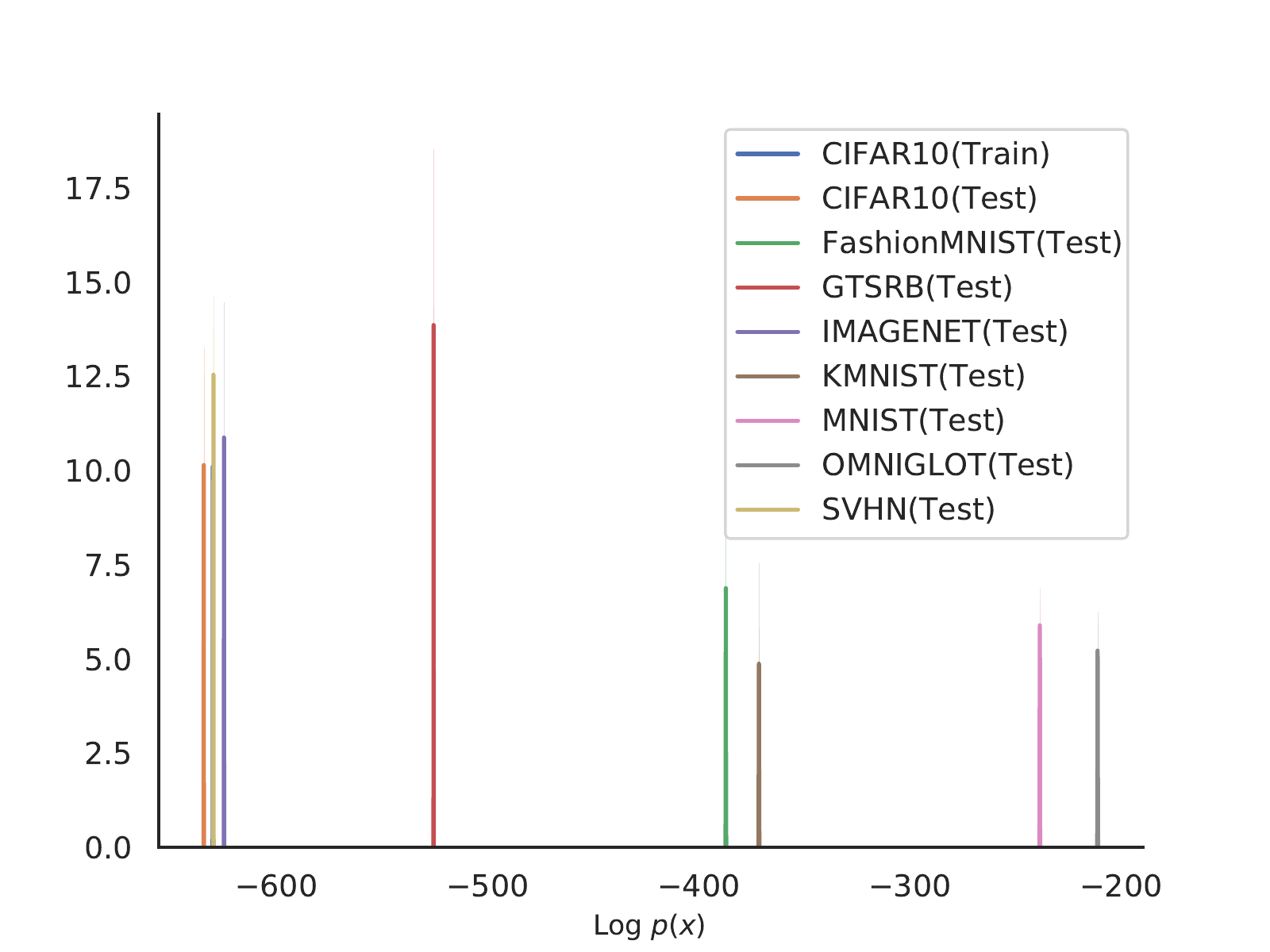}} 
\subfigure[Trained on SVHN]{\includegraphics[width=0.45\textwidth]{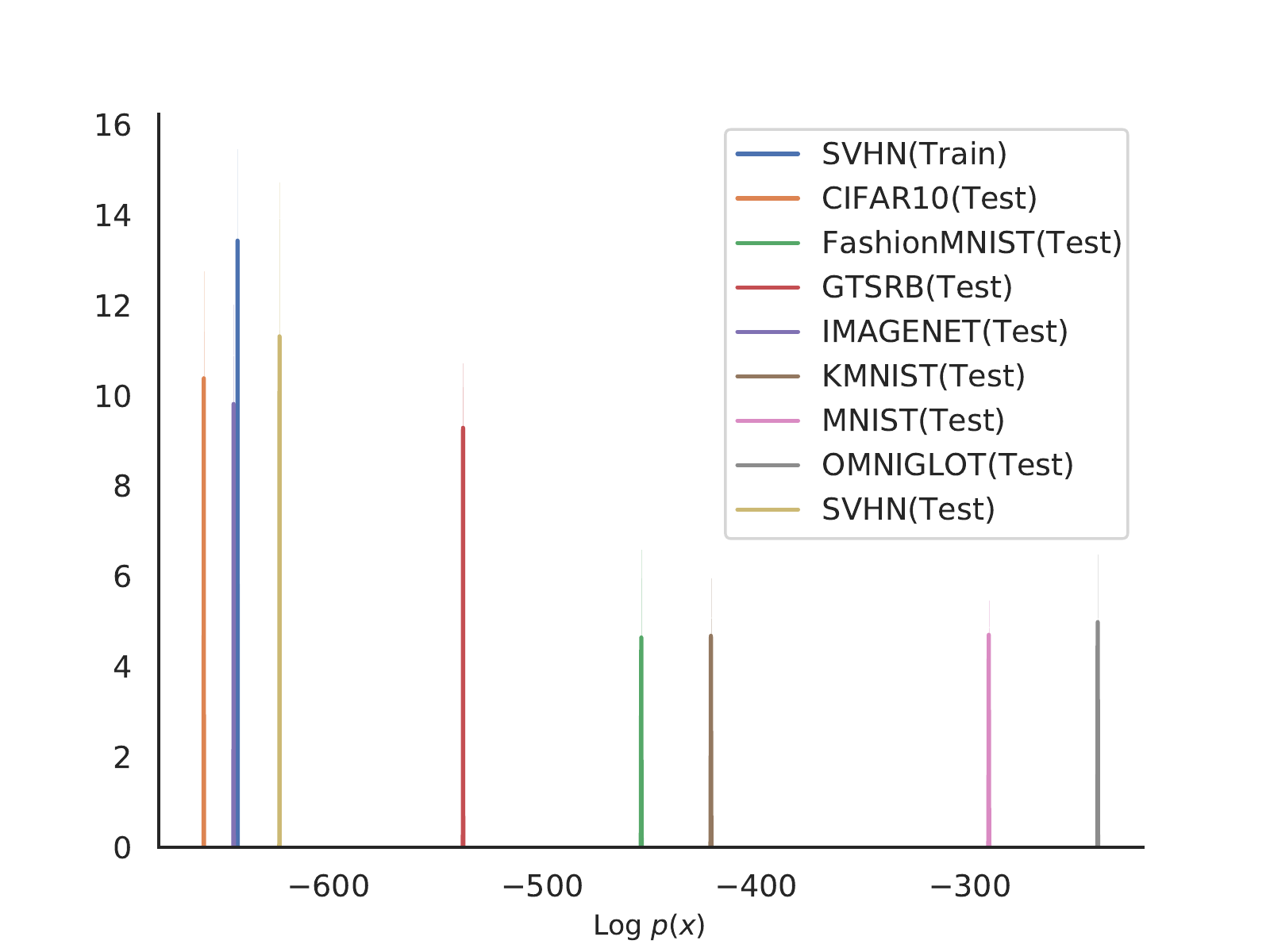}}\\
\subfigure[Trained on FashionMNIST]{\includegraphics[width=0.45\textwidth]{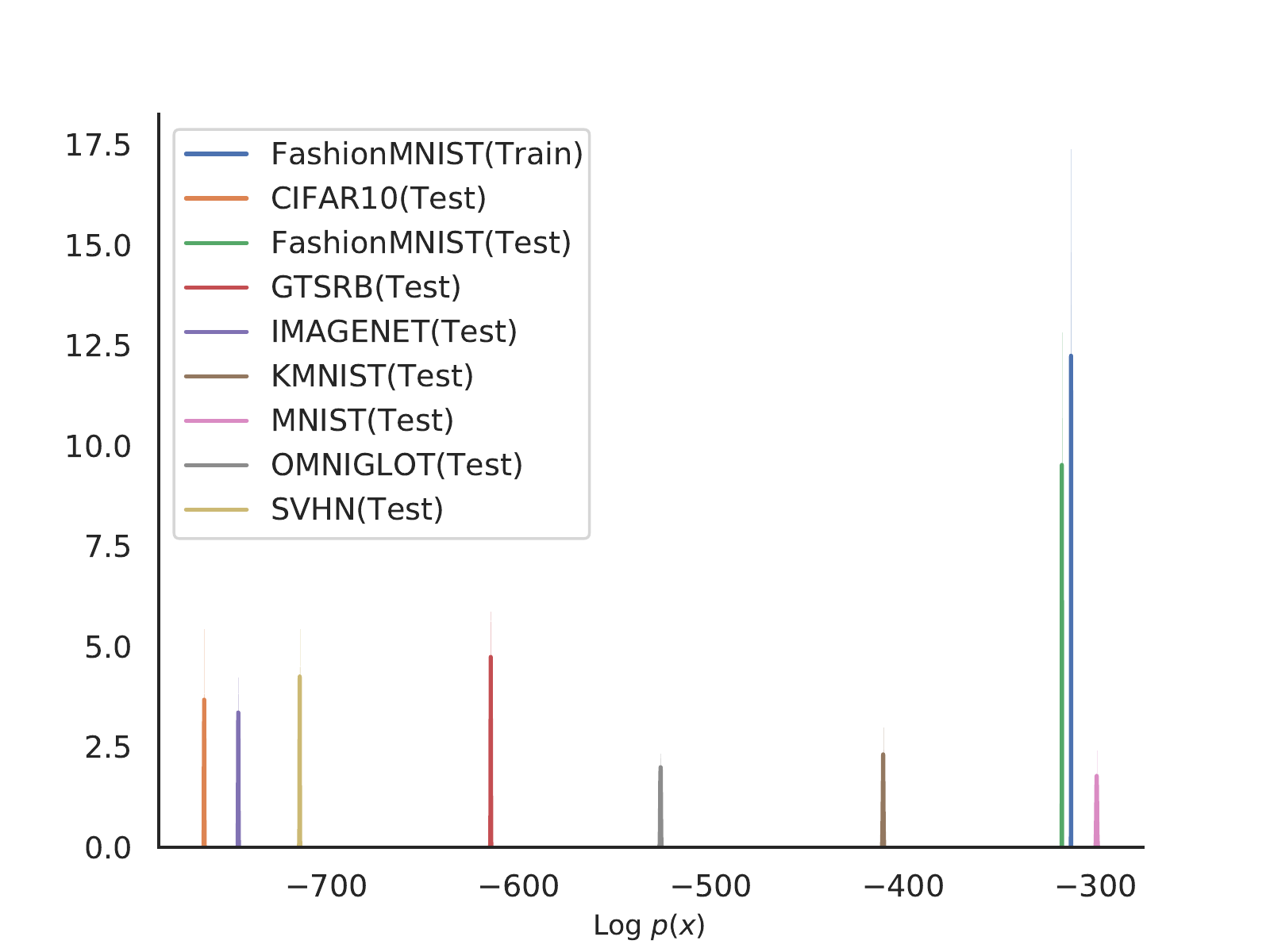}} 
\subfigure[Trained on OMNIGLOT]{\includegraphics[width=0.45\textwidth]{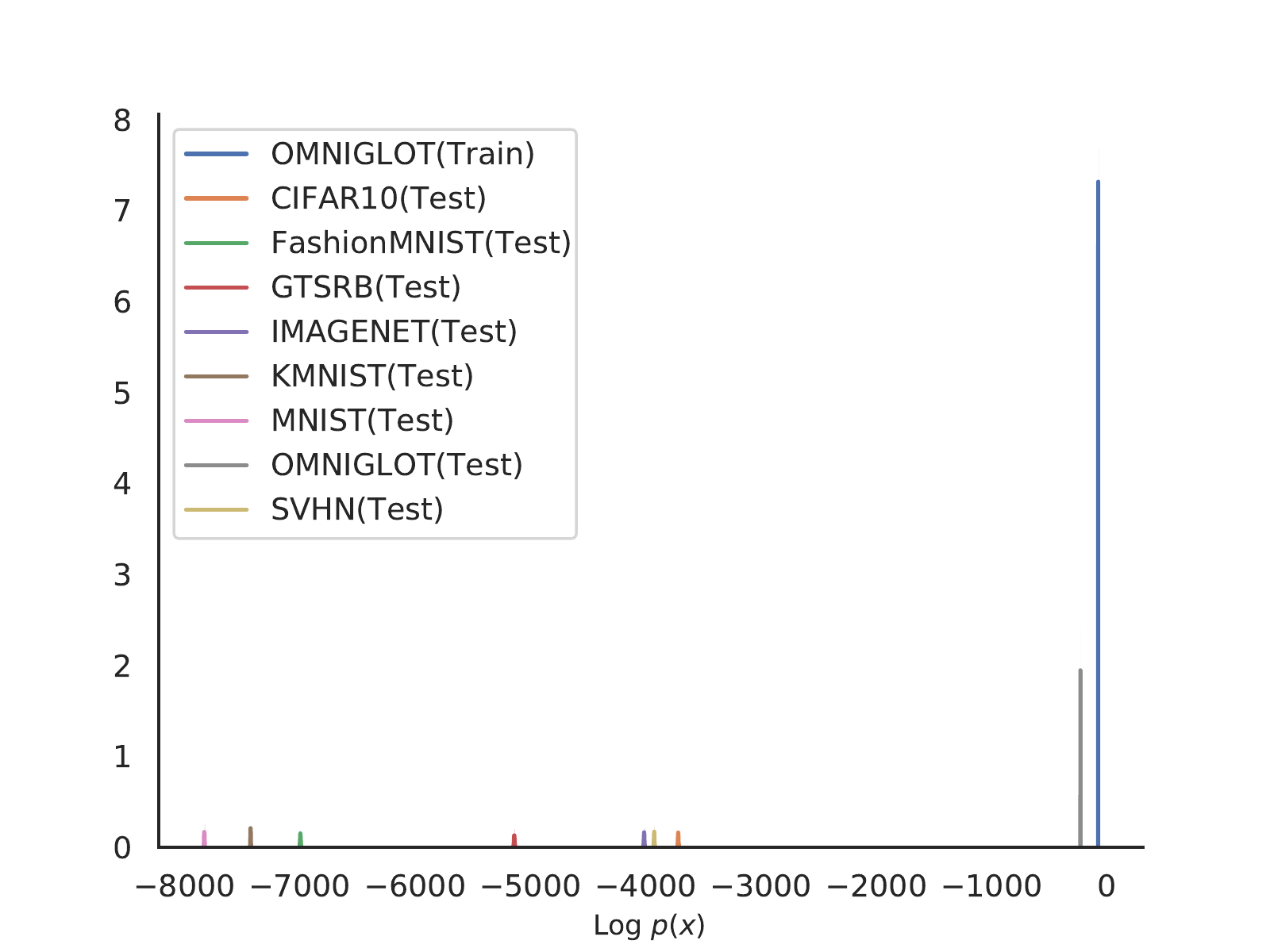}} 

\caption[]{ Histogram of log-likelihoods from a VAE model trained on CIFAR10, SVHN, FashionMNIST, and OMNIGLO. (see similar results in \cite{nalisnick2018deep,choi2018generative,serra2019input}). Other added results are shown in Figure \ref{fig_trn_likelihood_appre} in Appendix.}
\label{fig_trn_test_likelihood}
\vspace{0.1in}
\end{figure}

Out-of-distribution (OOD) detection is a crucial issue for machine learning (ML) security, which usually arises in many application scenarios, such as medical diagnosis and credit card fraud detection. There is a widely held view that likelihood-based generative models have strong robustness to the OOD inputs~\cite{bishop1994novelty, Blei2017panel}. Based on this opinion, a well-calibrated generative model can be a good detector by assigning higher likelihoods to the samples from the in-distribution (ID) dataset than OOD dataset. Hence, the deep generative models are generally considered as reliable for anomaly detection tasks~\citep{chalapathy2018group,xu2018unsupervised,ostrovski2017count}. However, recent works~\cite{nalisnick2018deep,hendrycks2018deep,choi2018generative,lee2017training,nalisnick2019detecting,huang2019out,maaloe2019biva} have reported the phenomenon that the density estimate of the deep generative model, in some cases, is not able to detect OOD inputs correctly. For instance,  VAEs~\cite{kingma2013auto,rezende2014stochastic} cannot identify images of common objects such as airplane, bird, cat, and dog (i.e., CIFAR10) from the OOD datasets (i.e., MNIST, FashionMNIST, KMNIST, and GTSRB), assigning higher likelihoods to the OOD samples when VAE is trained on CIFAR10 (Shown in \textbf{Figure~\ref{fig_trn_test_likelihood}a}). These findings conflict with the previous OOD detection method proposed by Bishop~\cite{bishop1994novelty}. To alleviate and resolve this issue, deep generative models are expected to understand the attributes of OOD data deeply and fully when utilizing density estimate detecting OOD samples.

\begin{figure}[H]
\centering
\subfigure[]{\includegraphics[width=0.5\textwidth]{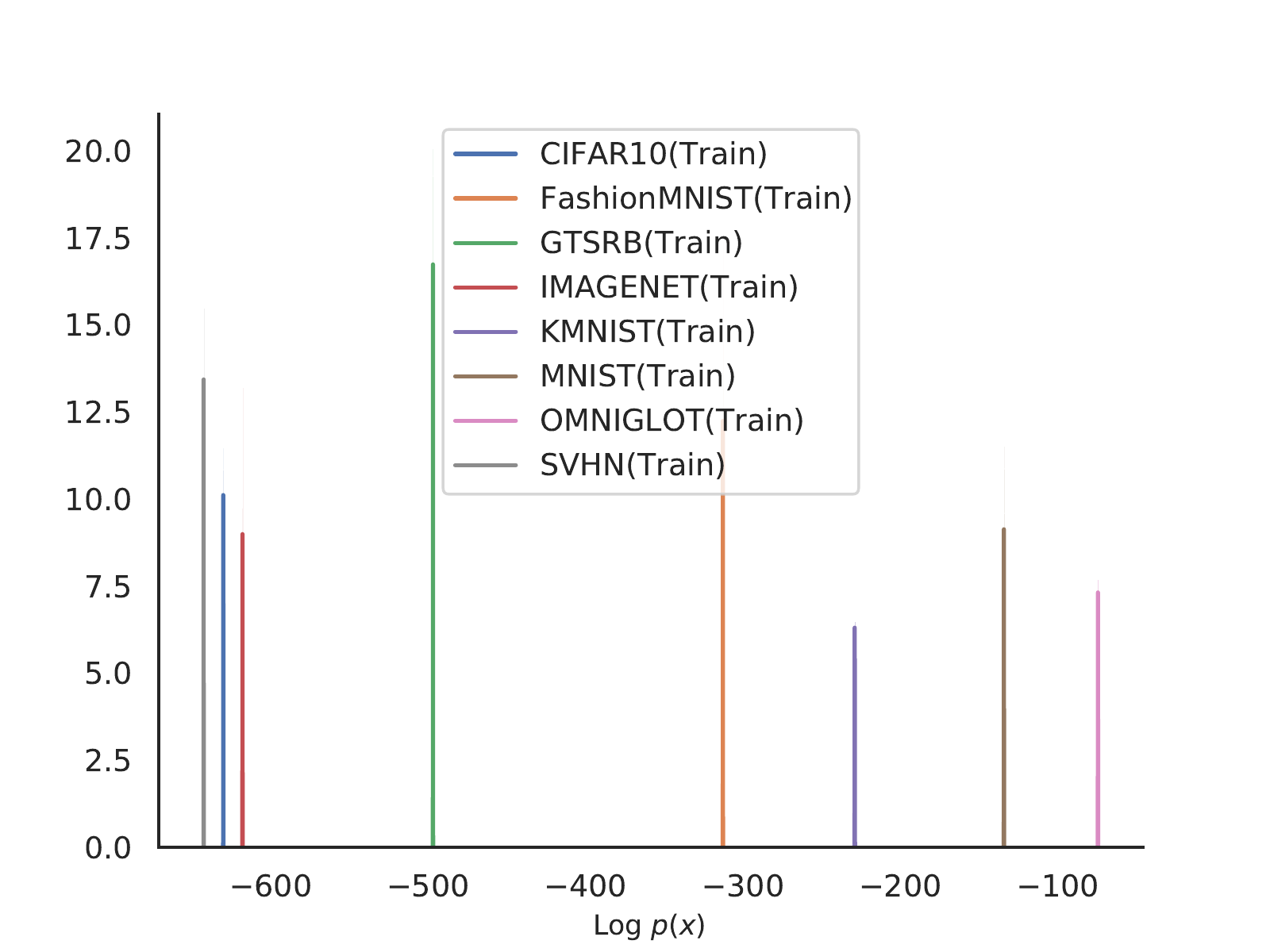}} 
\subfigure[]{\includegraphics[width=0.45\textwidth]{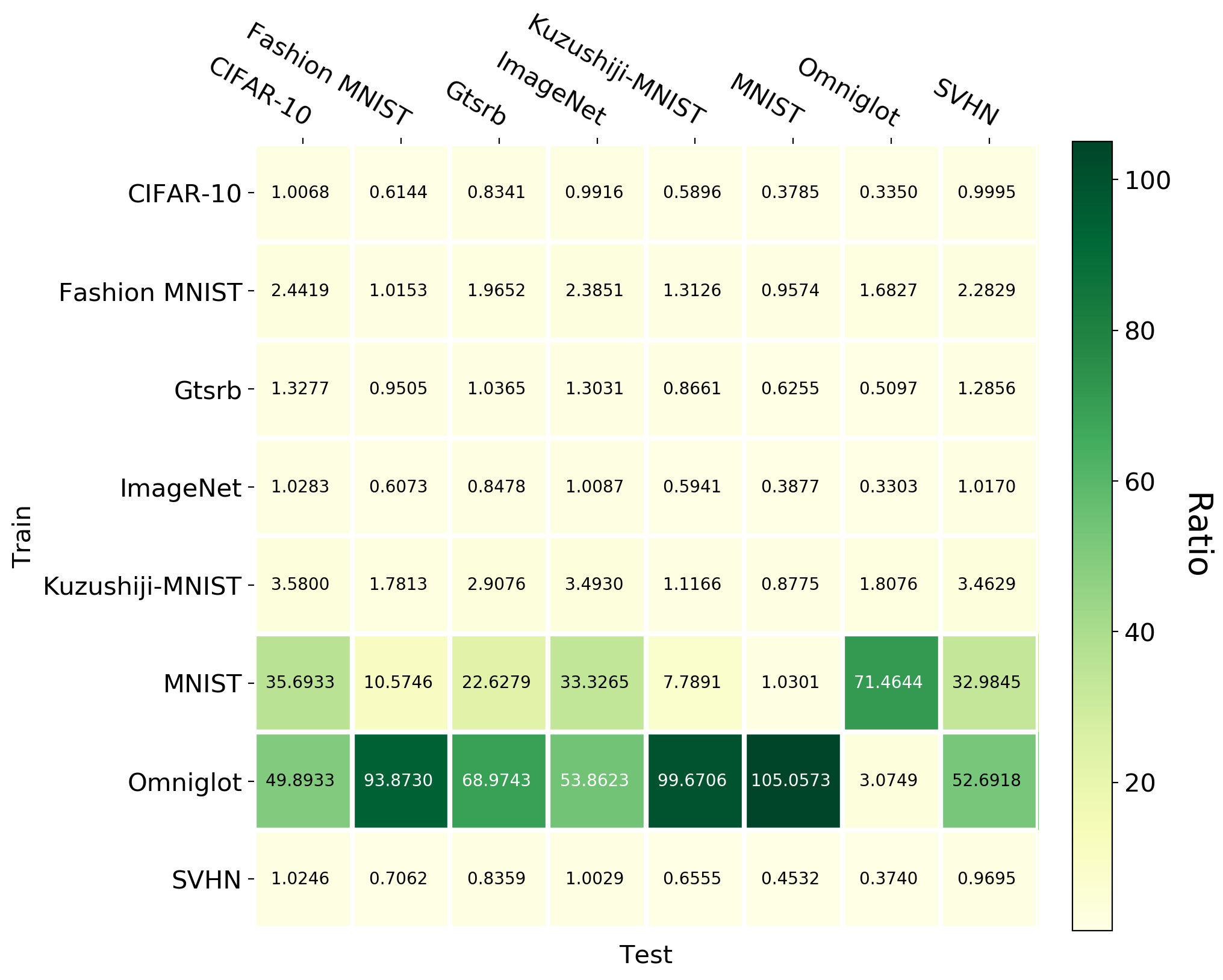}}\\
\caption[]{(a) Histogram of  log-likelihoods of VAE trained on Cifar10, FashionMnist, GTSRB, IMGAENET, KMNIST(Kuzushiji-MNIST), OMNIGLOT, and SVHN, respectively. (b) Likelihood Ratios of training and testing samples (\textit{the higher is better}). The Likelihood Ratios $<$ 1 is represented by  the likelihood of testing simple higher than training samples  }
\label{fig_trn_likelihood ratio}
\vspace{0.2in}
\end{figure}

A variety of works have emerged attempting to solve this problem. For instance, ~\cite{serra2019input} demonstrated that the input complexity affects greatly the density estimate of deep generative models by designing controlled experiments on Glow~\cite{kingma2018glow} model with different levels of image complexity. Similar qualitative results of VAEs are obtained in our experiments (\textbf{Figure~\ref{fig_trn_test_likelihood}}). Also, we computed the likelihoods of training samples from Cifar10, FashionMnist, GTSRB, IMGAENET, KMNIST, OMNIGLOT, and SVHN (Shown in  \textbf{Figure~\ref{fig_trn_likelihood ratio}a}). We find that the simple samples (KMNIST, OMNIGLOT, and MNIST) trained by VAEs with high likelihoods can assign lower likelihoods to the complex test samples (CIFAR10, SVHN, and IMGAENET) to detect OOD samples because the likelihood of complex samples trained on VAEs is smaller than the likelihood of simple samples trained on VAEs. In contrast, the VAEs trained on complex samples with a low likelihood usually give a higher likelihood for the simple test samples but identify them as ID samples (\textbf{Figure~\ref{fig_trn_likelihood ratio}b}). 
Inspired by this intriguing finding, here we propose a method that feeds the external dataset (called the simple dataset) as inputs while training VAEs on the training dataset (called the basic dataset), which is more straightforward than training VAE on the basic dataset. In this manner, VAEs can learn the features from two data distributions, assigning a higher likelihood for the basic dataset than the simple dataset. And the density estimate of VAEs can be used for detecting OOD samples.

\begin{figure*}[t]
\centering
\subfigure[]{\includegraphics[width=0.45\textwidth]{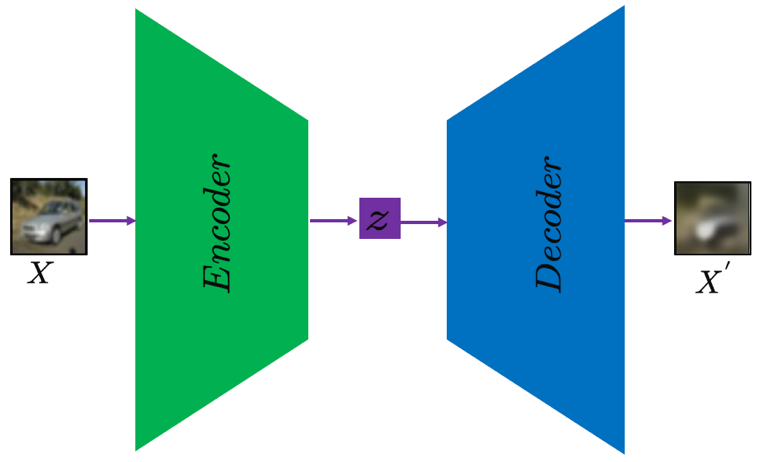}} 
\subfigure[]{\includegraphics[width=0.45\textwidth]{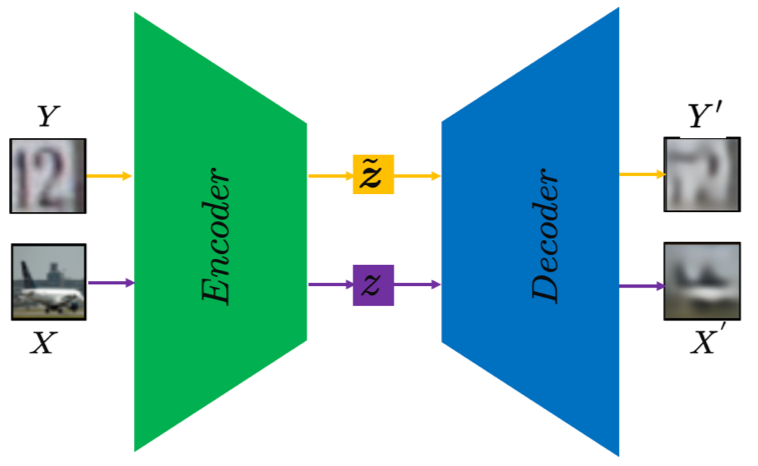}}\\
\subfigure[]{\includegraphics[width=0.45\textwidth]{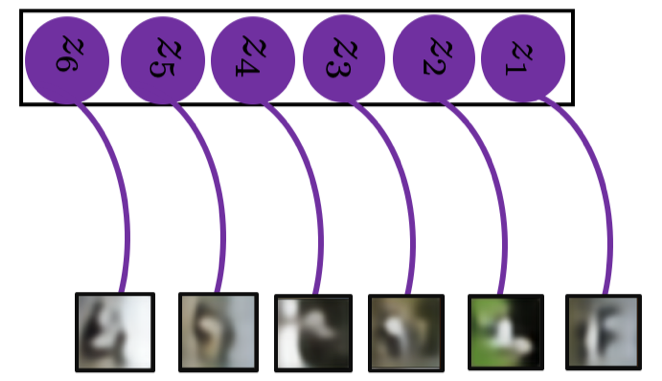}} 
\subfigure[]{\includegraphics[width=0.45\textwidth]{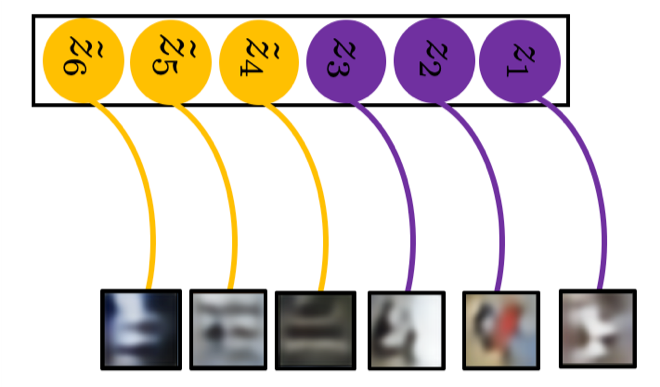}} 

\caption{Overview framework of standard VAE $(\textbf{a})$ and BPVAE$(\textbf{b})$. The encoder and decoder are represented by a green and blue trapezoid, respectively. The purple and yellow squares denote the latent space of baisc and simple data, respectively. Compared the generator of standard VAE $(\textbf{c})$ and BPVAE$(\textbf{d})$. The latent space of standard VAE only learn and capture basic, while the latent space of BPVAE cover the features of basic and simple data concurrently.}
\label{fig1_1}
\end{figure*}

In this work, we introduce the Bigeminal Priors Variational auto-encoder (BPVAE) (\textbf{Figure~\ref{fig1_1}}), an advanced extension of VAEs with two independent latent priors that belong to the basic and simple datasets, respectively. To build this hybrid model with an effective synergetic mode, two tricky problems arise. The first one is that how to choose the simple dataset and two priors for BPVAE. Due to a lot of other candidate datasets different from the basic dataset, it is hard to find the most appropriate candidate. here we firstly select a dataset randomly as a candidate and then train VAEs on the basic and candidate datasets, respectively. By comparing the likelihoods of samples from basic dataset with that of other candidate datasets, we can choose the right candidate dataset with higher likelihoods than the basic dataset as the simple dataset (\textbf{Figure~\ref{fig_trn_likelihood ratio}a}). For example, take GTSRB as the basic dataset, then FashionMNIST, MNIST, and KMNIST can be regarded as the simple datasets, while CIFAR10, IMAGENET, and SVHN cannot be the simple datasets.
On the other hand, there are plenty of candidate priors (e.g., standard normal distribution prior, Gaussian mixture prior~\cite{Dilokthanakul2016DeepUC}, Vamp Prior~\cite{tomczak2017vae}, Resampled Prior ~\cite{bauer2018resampled}, Reference prior~\cite{Bernardo1979ReferencePD,Berger2009THEFD,Nalisnick2017LearningAO}) for BPVAE. How to combine adaptive priors remains a vital step. Note that the BPVAE has two priors in the latent space. A good prior to basic dataset is expected to carry the pivotal features of the basic dataset, which is called the basic prior (b-prior for short). And a good prior to the simple dataset (s-prior for short) should cover the core features of the simple dataset and follow a distribution different from that of the basic dataset. Therefore, BPVAE with b-prior assigns a lower likelihood for the simple dataset. Overall, the uncertainty of b-prior is higher than s-prior, for the complexity and uncertainty of the datasets are positively correlated.

\section{Related work}
Various works ~\cite{nalisnick2018deep,choi2018generative,hendrycks2018deep,lee2017training} have reported that the deep generative models are not able to correctly detect OOD samples until the models have an excellent understanding of OOD inputs.
~\cite{maaloe2019biva} indicated that Bidirectional-Inference Variational Auto-encoder (BIVA) with the multiple latent layers could capture the high-level semantic information of the data, with a better understanding of OOD representation. However, the standard VAEs with one latent layer have a poor performance for anomaly detection. 
~\cite{choi2018generative} proposed an algorithm which takes the ensembles of generative models to estimate the Watanabe-Akaike Information Criterion (WAIC)~\cite{watanabe2010asymptotic} as a metric for outliers. 
~\cite{hendrycks2018deep} showed that robustness and uncertainty~\cite{malinin2018predictive,hafner2018reliable} concerning to the outlier exposure (OE) can be improved by training the model with OOD data, which can improve model calibration and several anomaly detection techniques were proposed accordingly.
~\cite{Ran2020DetectingOS} proposed an improved noise contrastive prior (INCP) to acquire a reliable uncertainty estimate for the standard VAEs. Patterns between OOD and ID inputs can be well captured and distinguished via VAEs with reliable uncertainty estimate.
~\cite{nalisnick2019detecting} proposed a statistical method to test the OOD inputs using a Monte Carlo estimate of empirical entropy, but their approach  is limited in the batches of inputs with the same type.~\cite{huang2019out} tried to use other generative models (i.e., neural rendering model (NRM)) for OOD detection, and they found the joint likelihoods of latent variables to be the most effective one for OOD detection.
~\cite{song2019unsupervised} demonstrate that OOD detection failure can induce sophisticated statistics based on the likelihoods of individual samples; they proposed a method that is in-batch dependencies for OOD detection.


\section{Approach}

\subsection{Variational autoencoder}
VAEs ~\cite{rezende2014stochastic,kingma2013auto} are a variety of latent variable models optimized by
the maximum marginal likelihood of an observation variable $ p(\bm{x}) $~\textbf{Figure \ref{fig1_1}a}. The marginal likelihood can be written as follows:

\begin{equation}
\label{e4}
\begin{aligned} 
\log p(\bm{x})= & \bm{E}_{\bm{z} \sim q_{\theta}(\bm{z} \mid \bm{x})}[\log p_{\phi}(\bm{x} \mid \bm{z})]-\bm{D}_{KL}[q_{\theta}(\bm{z} \mid \bm{x}) \| p(\bm{z})]\\
&+ \bm{D}_{KL}[q_{\theta}(\bm{z} \mid \bm{x}) \| p(\bm{z} \mid \bm{x})],
\end{aligned}
\end{equation}

where $p(\bm z)$ and $p(\bm{z} \mid \bm{x})$ are the prior by using a standard normal distribution and the true posterior respectively. $q_{\theta}(\bm{z} \mid \bm{x})$ is the variational posterior (encoder) by employing a Guassian distribution, and $p_{\phi}(\bm{x} \mid \bm{z})$ is the generative model (decoder) by using a Bernoulli distribution. Both are modeled by a neural network with their parameter  $\theta$, $\phi$, respectively. Thus, we train VAE with training samples to maximize the following objective variational evidence lower bound (ELBO):

\begin{equation}
\label{e5}
\begin{aligned} 
\mathcal{L}_{}(\phi, \theta)=  \bm{E}_{\bm{z} \sim q_{\theta}(\bm{z} \mid \bm{x})}[\log p_{\phi}(\bm{x} \mid \bm{z})]-\bm{D}_{KL}[q_{\theta}(\bm{z} \mid \bm{x}) \| p(\bm z)]
\\
\end{aligned}
\end{equation}

where $q_{\theta}(\bm{z} \mid \bm{x})$  and $q_{\theta}(\bm{\tilde z} \mid \bm{\tilde x})$ are variational posteriors for matching the true posteriors ($p_{}(\bm{z} \mid \bm{x})$  and $p_{}(\bm{\tilde z} \mid \bm{\tilde x})$) which are given by $ \bm{\tilde x}$ and $\bm{x}$ respectively.
For a given dataset, the marginal likelihood is a constant. From Eq.~\ref{e5} and Eq.~\ref{e4}, we get

\begin{equation}
\label{e6}
\begin{aligned} 
\log p(\bm{x})  \geq \mathcal{L}_{}(\phi, \theta) 
\end{aligned}
\end{equation}

Assuming variational posterior has arbitrarily high-capacity for modeling, $q_{\theta}(\bm{z} \mid \bm{x})$  approximates intractably $p(\bm{z} \mid \bm{x})$ and the KL-divergence between $q_{\theta}(\bm{z} \mid \bm{x})$ and  $p(\bm{z} \mid \bm{x})$ will be zero. The $\mathcal{L}_{}(\phi, \theta) $ can be replaced by $\log p(\bm{x})$.

\subsection{BIGEMINAL PRIORS Variational autoencoder}

The BPVAE consists of an encoder, a decoder, and two priors (b-prior and s-prior)~\textbf{Figure \ref{fig1_1}b}, which is trained on both the basic dataset learned by b-prior and the simple dataset learned by s-prior. Specifically, the uncertainty of b-prior is higher than s-prior due to the positive correlation between the complexity and uncertainty of the dataset. We assume that both the b-prior and s-prior belong to normal distribution. And we use the variance of a normal distribution to represent the uncertainty level. The priors are formulated as followings,

\begin{equation}
\begin{aligned} 
p_{b}(z) & \sim \mathcal{N}(z \mid \mu_{z}, \sigma^{2}_{z}I) \\ p_{s}(\tilde z ) & \sim \mathcal{N}\left(\tilde{z} \mid \mu_{\tilde z} , \sigma^{2}_{\tilde z} I\right)
\end{aligned}
\end{equation}

where the mean value $\mu_{z}$=$\mu_{\tilde{z}}$= $0$. And the variances $\sigma_{z} $ and $ \sigma_{\tilde z}$ are hyper-parameters determining the uncertainty of b-prior and s-prior.  $\sigma_{ z} $ is always set to be greater than $ \sigma_{\tilde z}$ so that b-prior has enough capacity to capture the basic dataset features. In this manner, BPVAE can capture the features of basic dataset, as well as of simple datatset. To facilitate the training implementation, we modified the loss function as follow:

\begin{equation}
\begin{aligned} \log p(\boldsymbol{x}) + \log p(\boldsymbol{y})=& \boldsymbol{E}_{\boldsymbol{z} \sim q_{\theta}(\boldsymbol{z} \mid \boldsymbol{x})}\left[\log p_{\phi}(\boldsymbol{x} \mid \boldsymbol{z})\right]-\boldsymbol{D}_{K L}\left[q_{\theta}(\boldsymbol{z} \mid \boldsymbol{x}) \| p_{b}(\boldsymbol{z})\right] \\ &+\boldsymbol{E}_{\boldsymbol{\tilde z} \sim q_{\theta}(\boldsymbol{\tilde z} \mid \boldsymbol{y})}\left[\log p_{\phi}(\boldsymbol{y} \mid \boldsymbol{\tilde z})\right]-\boldsymbol{D}_{K L}\left[q_{\theta}(\boldsymbol{\tilde z} \mid \boldsymbol{y}) \| p_{b}(\boldsymbol{\tilde z})\right]
\end{aligned}
\end{equation}

where $q_{\theta}(\tilde{\boldsymbol{z}}\mid {\boldsymbol{y}})$  and $q_{\theta}(\boldsymbol{z} \mid \boldsymbol{x})$ are the variational posterior for the simple and basic dataset,   $q_{\theta}(\tilde{\boldsymbol{z}}\mid \boldsymbol{y})$ and $q_{\theta}(\boldsymbol{z} \mid \boldsymbol{x})$ are the decoder for the simple and basic data, and which are modeled by a neural network with their parameters $\theta$ and $\phi$, respectively.

\section{RESULTS}

\begin{figure}[t]
\centering
\subfigure[Test on MNIST]{\includegraphics[width=0.8\textwidth]{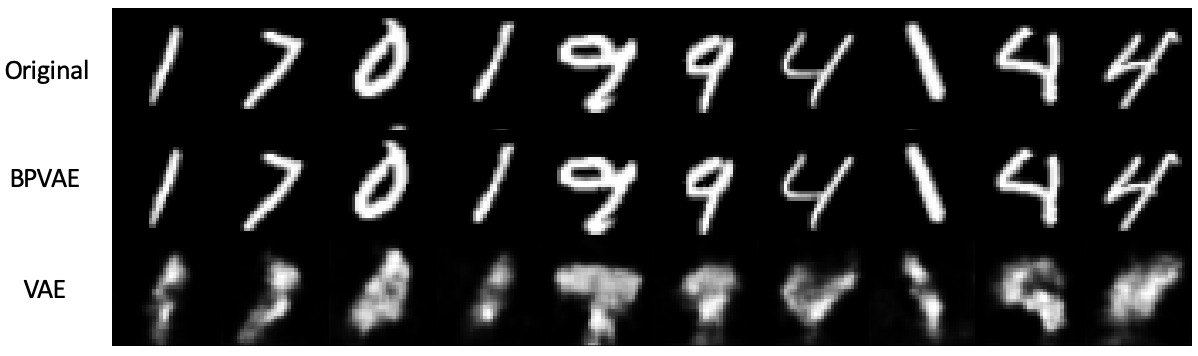}} 
\subfigure[Test on CIFAR10]{\includegraphics[width=0.8\textwidth]{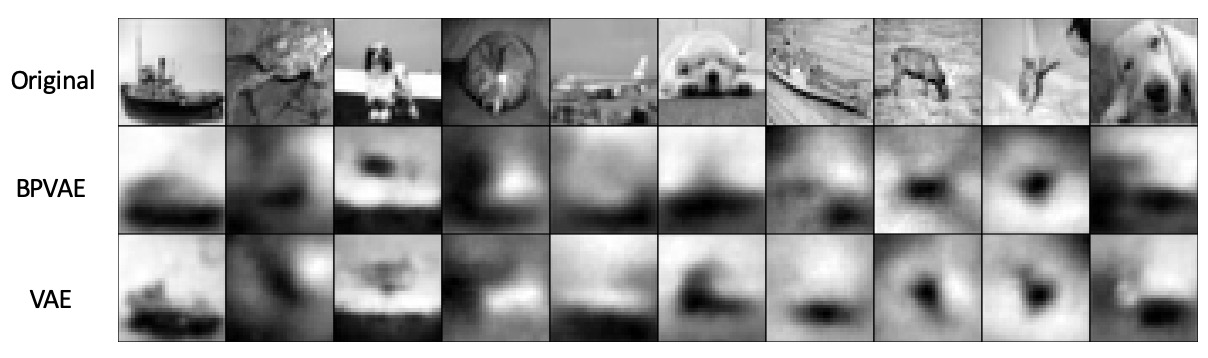}}
\caption[]{Reconstruction performance for MNIST and CIFAR10 by VAEs and BPVAEs. Here CIFAR10 is used as basic dataset and MNIST is used as simple dataset.}
\label{reconstruction}
\vspace{0.1in}
\end{figure}

\subsection{Does BPVAE know what it doesn't know?}
To investigate whether VAEs have a good understanding of the distribution of training data, we carry out reconstruction experiments under multiple conditions. Despite the variety of choices for settings of reconstruction experiment, here we take the following setting as an example: CIFAR10 as the basic dataset for training and MNIST as the simple dataset. After training VAEs and BPVAEs separately, we generated reconstructed images during the inference process. As shown in~\textbf{Figure~\ref{reconstruction}}, we visualized the results of standard VAEs and our proposed BPVAEs in comparison. It is evident that BPVAEs obtain much better performance than standard VAEs on MNIST, while these two models achieve comparable results on CIFAR10. The great performance of BPVAEs on MNIST can be attributed to the effective capacity of the extra introduced s-priors, which can assist BPVAEs of capturing external feature representation for the data from simple dataset, in which case VAEs failed due to lack of various latent priors with strong capacity.


Besides, we evaluate the reconstruction effects quantitatively by using MSE (Mean Squared Error), PSNR (Peak Signal to Noise Ratio) and SSIM (Structural Similarity)~\cite{Hor2010Image}. Note that as for PSNR, the value lower is better, and as for PSNR and SSIM, the value higher is better.

Results for comprehensive comparisons between two models are presented in~\textbf{Table \ref{table1}}. The tables demonstrate that BPVAEs can obtain much better performance than standard VAEs no matter it is evaluated by MSE, PSNR or SSIM, meanwhile retaining the comparable capacity to capture and reconstruct data from basic dataset, which is consistent with the aforementioned descriptions from qualitative observation.


\begin{table}[]
\caption{Evaluation on the basic dataset and the simple dataset}
\label{table1}
\centering
\begin{tabular}{lllll}
\toprule
& Method       & MSE    & PSNR   & SSIM  \\
\midrule
\multirow{2}{*}{Evaluation on basic dataset} & BPVAE & 0.017  & 18.250 & 0.544 \\
& VAE   & 0.016  & 18.282 & 0.543 \\
\midrule
\multirow{2}{*}{Evaluation on simple dataset} & BPVAE & 0.007  & 22.392 & 0.909 \\
& VAE   & 0.0346 & 14.831 & 0.601\\
\bottomrule
\end{tabular}
\end{table}

\subsection{Analysis}
\begin{figure}[H]
\centering

\subfigure[Trained on CIFAR10 and FahionMINST]{\includegraphics[width=0.45\textwidth]{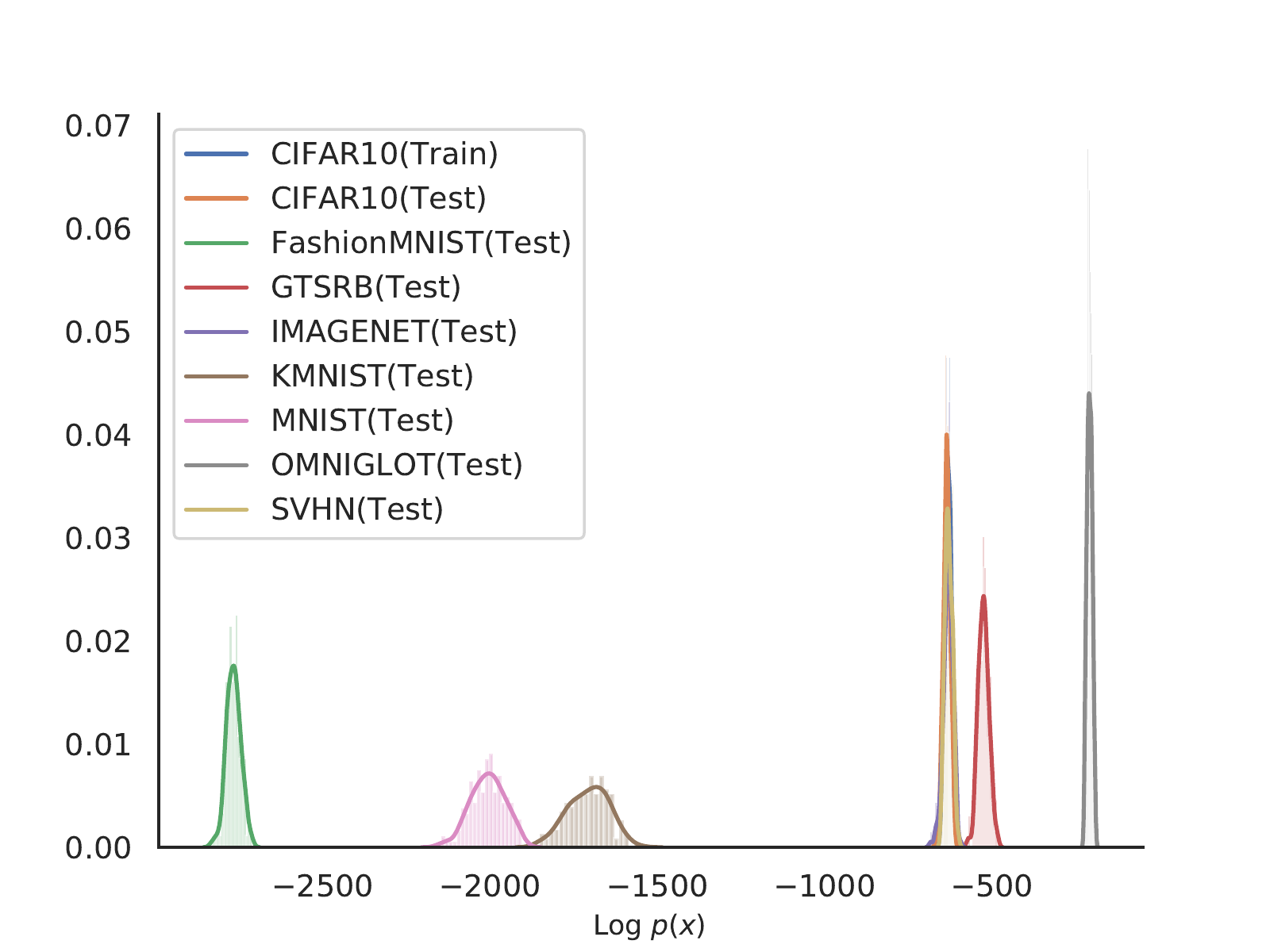}} 
\subfigure[Trained on CIFAR10 and IMAGENET]{\includegraphics[width=0.45\textwidth]{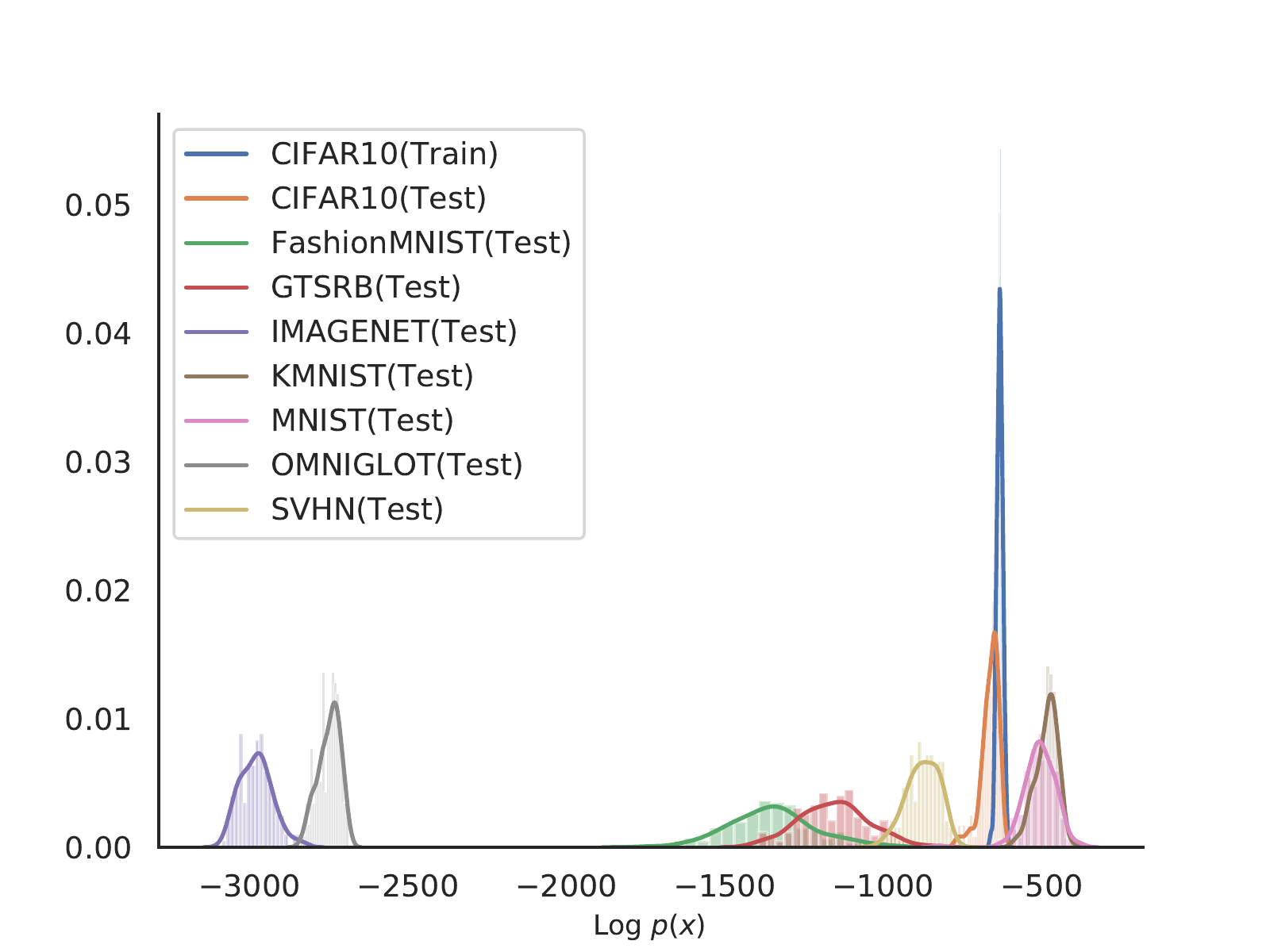}} 
\subfigure[Trained on CIFAR10 and KMNIST]{\includegraphics[width=0.45\textwidth]{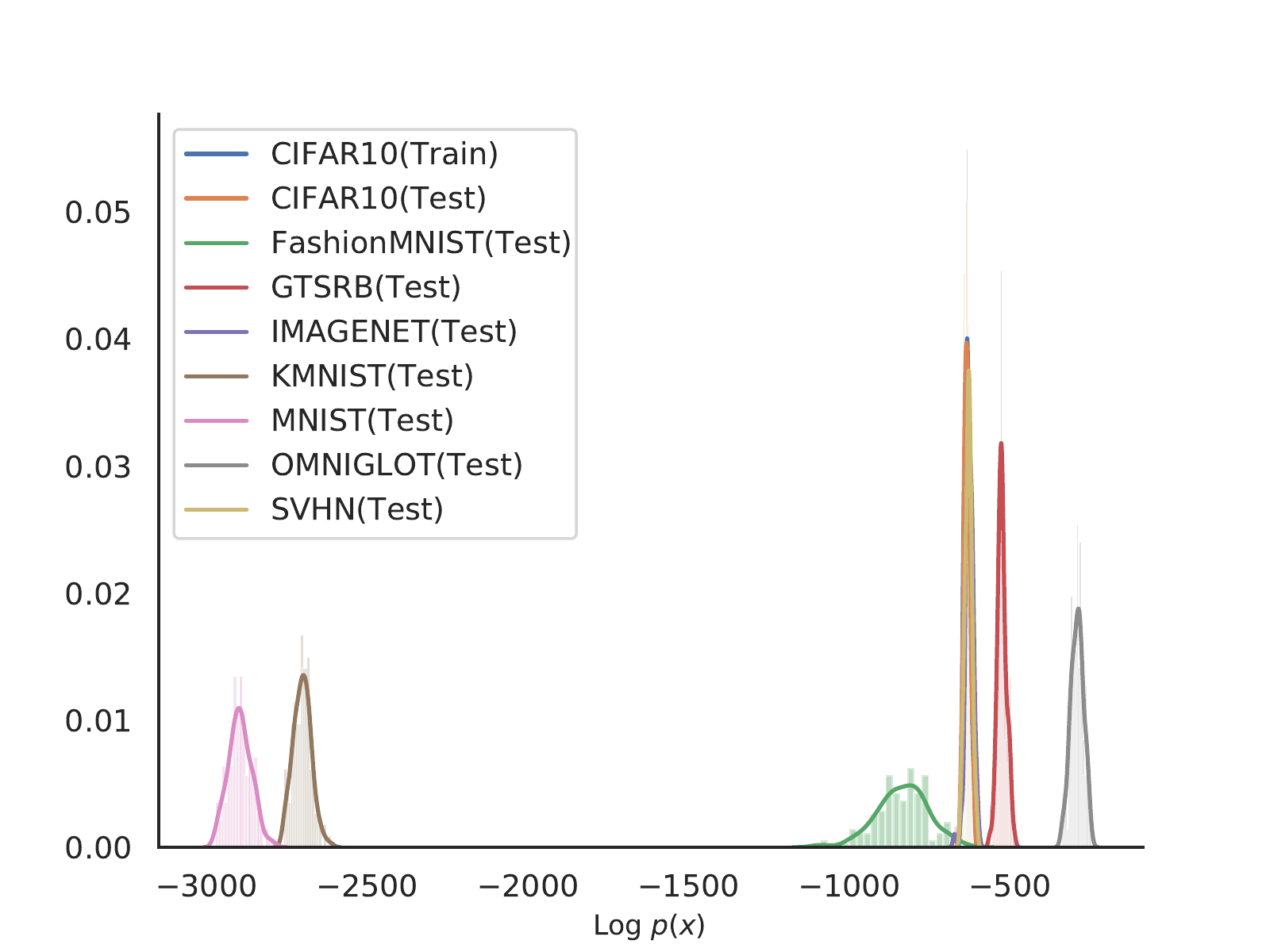}}
\subfigure[Trained on CIFAR10 and KMNIST]{\includegraphics[width=0.45\textwidth]{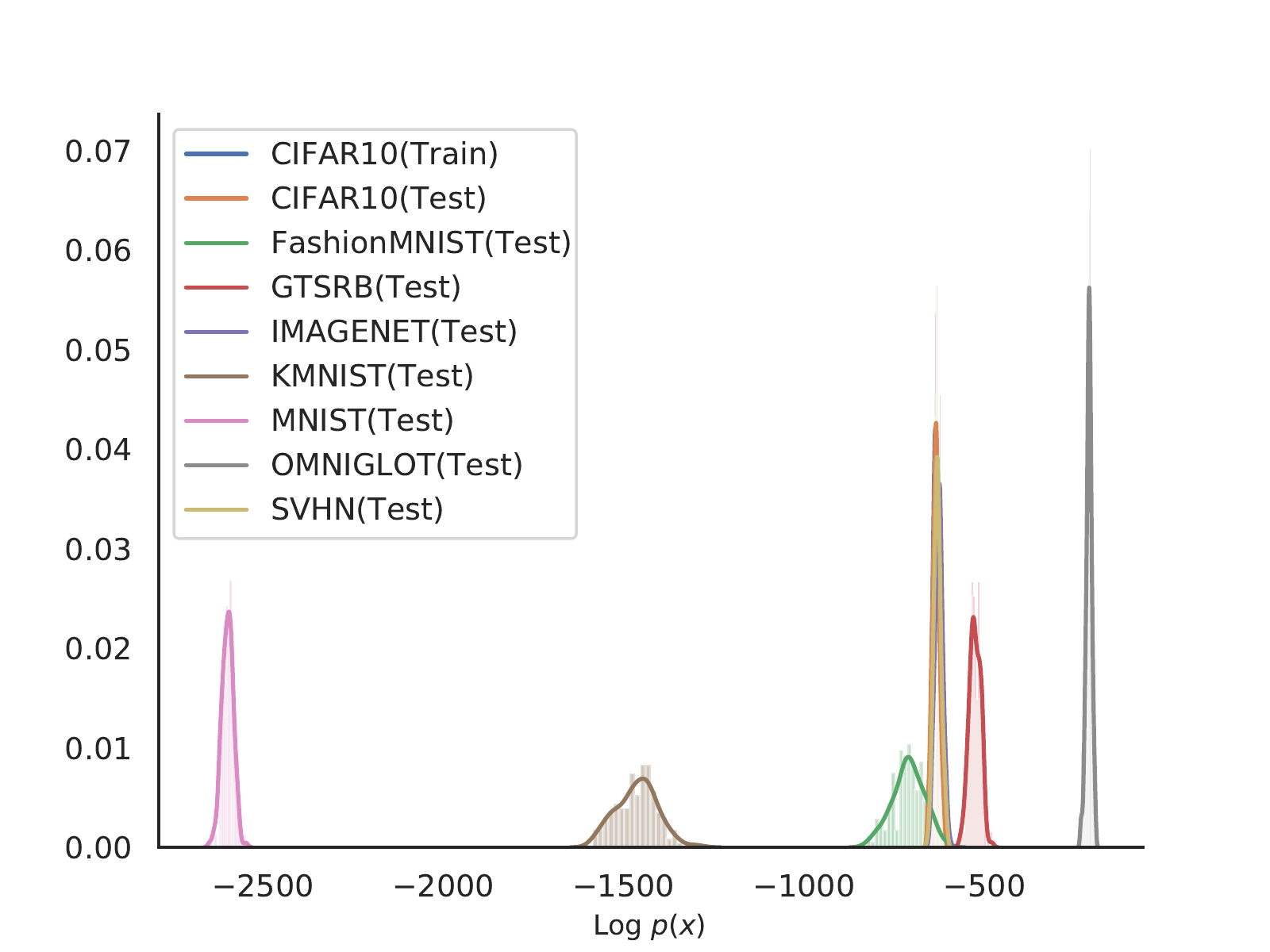}} 
\caption[]{Histogram of log-likelihoods from VAEs model, which are trained on different groups of datasets.}
\label{analysis}
\end{figure}

To explore the internal mechanism of BPVAEs, we perform some comparison experiments with multiple OOD testing samples from the different data distribution. As described in ~\textbf{Figure \ref{analysis}}, we train the BPVAE model on CIAFR10 (basic dataset), meanwhile with other datasets as simple dataset. Take~\textbf{Figure \ref{analysis}a} as an example, after the training process on CIFAR10 and FashionMNIST, BPVAEs can output lower likelihoods for most low-complexity datasets (such as MNIST, SVHN, etc.), avoiding the excessive-high likelihood problem for the simple data. This is because VAEs with the hybrid prior mode are equipped with a higher capacity of representation and therefore can shift the input distribution towards the lower-likelihood direction. On the contrary, for GTSRB and KMNIST in this case, BPVAEs fail to transform their distribution into representation in lower space of data distribution.  ~\textbf{Figure \ref{analysis}b,c,d} present similar phenomenon. This illustrates that although adding extra prior can indeed facilitate the VAEs' robustness and representation capacity, alleviating OOD problem by shifting data distribution of low-complexity dataset, but the distribution scale where it can cover is not infinite, which usually lies in the nearby neighborhood from the data distribution captured by b-prior and s-prior.

\begin{figure}[H]
\centering
\subfigure[]{\includegraphics[width=0.45\textwidth]{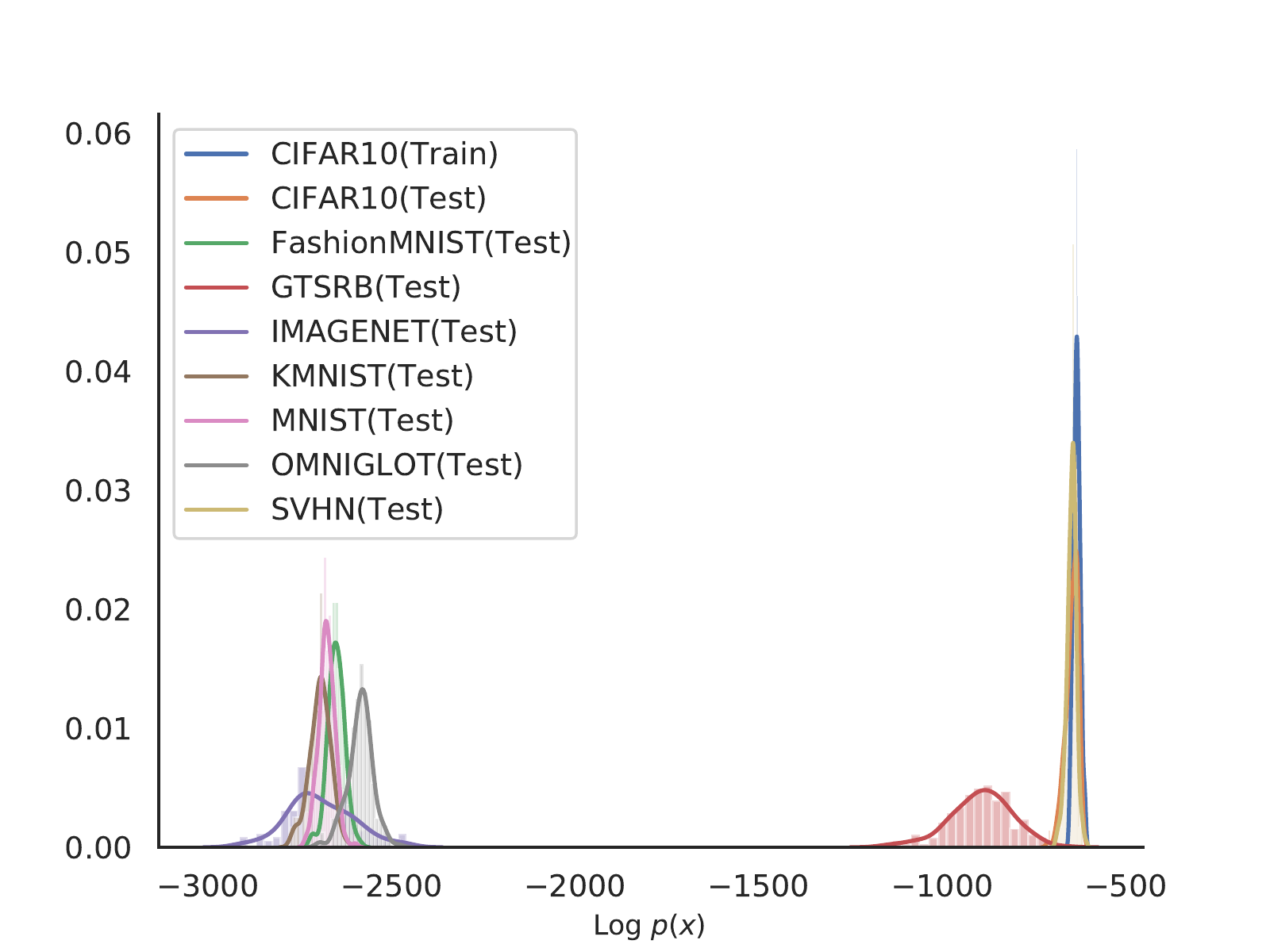}} 
\subfigure[]{\includegraphics[width=0.45\textwidth]{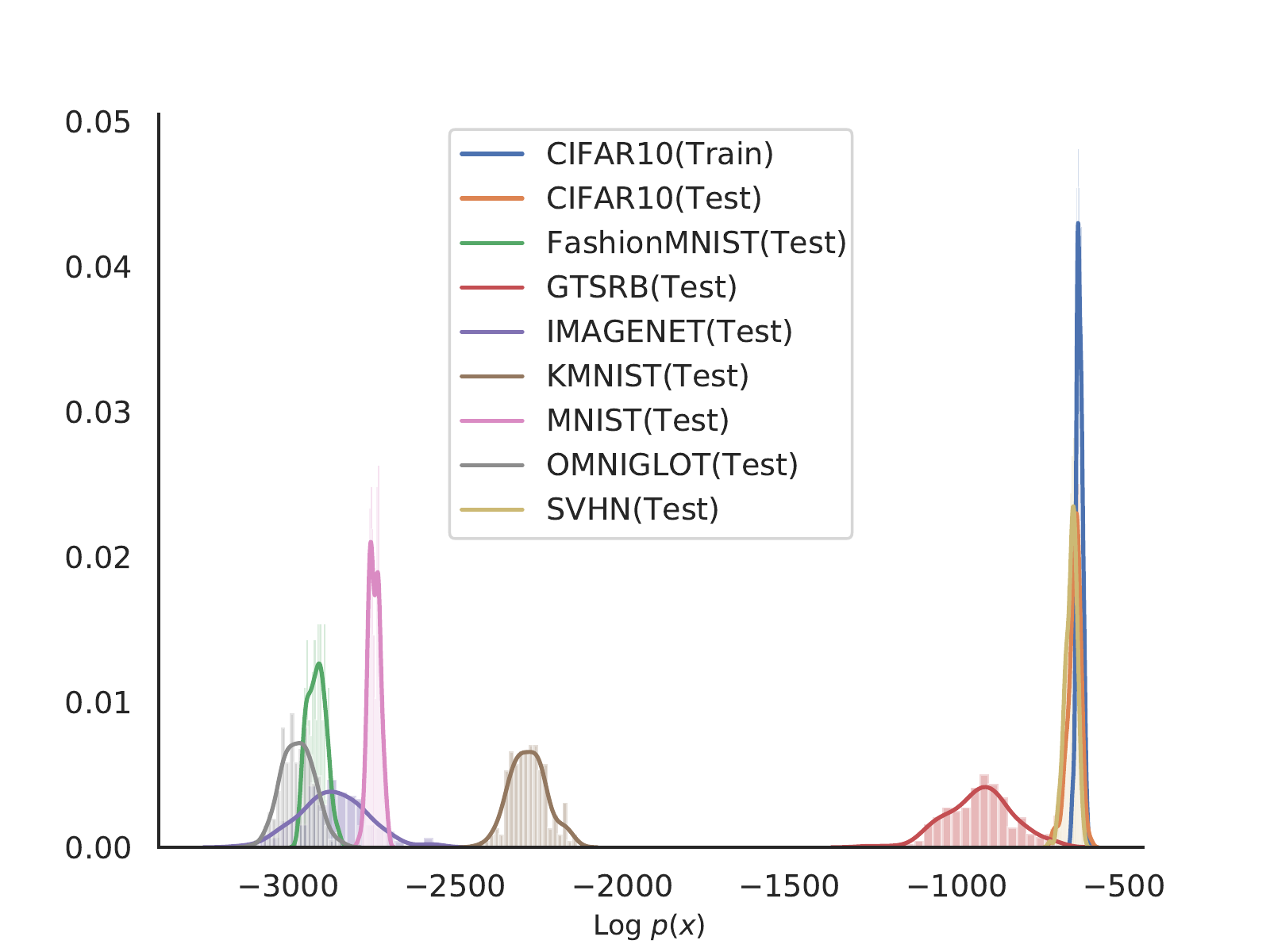}} \\
\caption[]{Histogram of log-likelihoods from VAEs model, which are trained on different groups of datasets. (a) Trained on CIFAR10(Basic), FahionMINST(simple), and KMINST(simple); (b) Trained on CIFAR10(Basic), FahionMINST(simple), and MINST(simple);}
\label{analysis2}
\vspace{0.2in}
\end{figure}

In order to alleviate the aforementioned limitation, we propose a more comprehensive approach to broaden its applicable scale. As \textbf{Figure \ref{analysis2}} presents, our model can cover all key representation and shift all the data distribution toward the lower-likelihood area, via combining multiple priors and training BPVAEs on a variety of selected datasets. This is presumably helpful for OOD detection as well, and we will show its performance on OOD detection tasks in the next section.

\subsection{OOD Detection}
\label{ood}
We perform OOD detection experiments on FashionMNIST and CIFAR10 datasets. For gray images, we train VAEs on FashionMNIST, train BPVAEs on FashionMNIST (basic dataset) and OMNIGLOT (simple dataset). And then we conduct OOD test with MNIST data as inputs. For RGB images, we train VAEs on CIFAR10, train BPVAEs on CIFAR10 (basic dataset) and GTSRB (simple dataset). And then we conduct OOD test with SVHN data as inputs. As depicted in Table \ref{sample-table0} and \ref{sample-table1}, our BPVAEs can achieve higher AUROC and AUPRC values then Standard VAEs, meanwhile surpassing other classical baselines. Overall, these comprehensive comparisons suggest that our proposed model is equipped with strong robustness and detection capability.

\begin{table}[H]
  \caption{AUROC and AUPRC for detecting OOD inputs using likelihoods of BPVAE, likelihood of VAE, and other baselines on FashionMNIST vs. MNIST datasets.}
  \label{sample-table0}
  \centering
  \begin{tabular}{lll}
    \toprule
    \cmidrule(r){1-2}
    Model     & AUROC    & AUPRC  \\
    \midrule
    BPVAE(ours)  & $\bm{1.000}$  & $\bm{1.000}$ \\
    Standard VAE & 0.012  & 0.113    \\
    Likelihood Ratio($\mu$, $\lambda$)~\cite{ren2019likelihood}    & 0.994 & 0.993      \\
    ODIN ~\cite{liang2017enhancing}    & 0.752       & 0.763  \\
    Mahalanobis distance~\cite{lee2018simple} & 0.942  & 0.928  \\
    Ensemble, 20 classifiers~\cite{lakshminarayanan2017simple} & 0.857 & 0.849  \\
    WAIC,5 models ~\cite{choi2018generative} & 0.221  & 0.401  \\
    \bottomrule
  \end{tabular}
\end{table}

\begin{table}[H]
  \caption{AUROC and AUPRC for detecting OOD inputs using likelihoods of BPVAE and VAE, and other baselines on CIFAR10 vs. SVHN datasets.}
  \label{sample-table1}
  \centering
  \begin{tabular}{lll}
    \toprule
    \cmidrule(r){1-2}
    Model     & AUROC    & AUPRC  \\
    \midrule
    BPVAE(ours)  & $\bm{1.000}$  & $\bm{1.000}$ \\
    Standard VAE & 0.037 & 0.214     \\
    Likelihood Ratio($\mu$, $\lambda$)~\cite{ren2019likelihood}     & 0.930 & 0.881  \\
    ODIN ~\cite{liang2017enhancing}    &  0.938        & 0.926  \\
    Mahalanobis distance~\cite{lee2018simple} &  0.728   & 0.711  \\
    Ensemble, 20 classifiers~\cite{lakshminarayanan2017simple} & 0.946 & 0.916  \\
    WAIC,5 models ~\cite{choi2018generative} & 0.146  & 0.363  \\
    \bottomrule
  \end{tabular}
\end{table}

\section{Discussion}
OOD problem has been increasingly gaining attention and interests, which remains an intriguing property and a challenging issue for likelihood-based generative models. In this work, we introduced external latent priors to assist VAEs in capturing more abstract representations for data which not belong to in-distribution. Through building an effective synergistic mode, VAEs can obtain powerful representation ability for different data from various datasets. In this manner, VAEs can be well-calibrated by shifting the likelihood distribution of data with simpler complexity to lower-likelihood intervals compared to basic dataset, in which way the high-likelihoods problem of OOD can be overcome to a large extent. Interestingly, we find there is a trivial trade-off when employing detection tasks, that is, even this method can alleviate OOD problem to a great extent, the likelihood interval scale which can be covered by bridging two latent priors is a little limited. Hence we introduce a hybrid VAE version with multiply latent priors, which can alleviate the trade-off greatly. Besides, we only impose the proposed approach on VAE model, designing the hybrid latent priors for other models like Glow, PixelCNN~\cite{van2016conditional} will be an interesting research topic. And we are expected to continue related exploration further. Overall, from a brand-new perspective, this work provides a potential way to tackle the OOD problem intertwined with VAEs.

{\small

\begin{thebibliography}{33}
\providecommand{\natexlab}[1]{#1}
\providecommand{\url}[1]{\texttt{#1}}
\expandafter\ifx\csname urlstyle\endcsname\relax
  \providecommand{\doi}[1]{doi: #1}\else
  \providecommand{\doi}{doi: \begingroup \urlstyle{rm}\Url}\fi

\bibitem[Bauer and Mnih(2018)]{bauer2018resampled}
M.~Bauer and A.~Mnih.
\newblock Resampled priors for variational autoencoders.
\newblock \emph{arXiv preprint arXiv:1810.11428}, 2018.

\bibitem[Berger et~al.(2009)Berger, Bernardo, and Sun]{Berger2009THEFD}
J.~O. Berger, J.~M. Bernardo, and D.~Sun.
\newblock The formal definition of reference priors.
\newblock \emph{Annals of Statistics}, 37:\penalty0 905--938, 2009.

\bibitem[Bernardo(1979)]{Bernardo1979ReferencePD}
J.~M. Bernardo.
\newblock Reference posterior distributions for bayesian inference.
\newblock 1979.

\bibitem[Bishop(1994)]{bishop1994novelty}
C.~M. Bishop.
\newblock Novelty detection and neural network validation.
\newblock \emph{IEE Proceedings-Vision, Image and Signal processing},
  141\penalty0 (4):\penalty0 217--222, 1994.

\bibitem[Blei et~al.(2017)Blei, Heller, Salimans, Welling, and
  Ghahramani]{Blei2017panel}
D.~Blei, K.~Heller, T.~Salimans, M.~Welling, and Z.~Ghahramani.
\newblock Panel discussion. advances in approximate bayesian inference.
\newblock \emph{NeurIPS Workshop}, 2017.

\bibitem[Chalapathy et~al.(2018)Chalapathy, Toth, and
  Chawla]{chalapathy2018group}
R.~Chalapathy, E.~Toth, and S.~Chawla.
\newblock Group anomaly detection using deep generative models.
\newblock In \emph{Joint European Conference on Machine Learning and Knowledge
  Discovery in Databases}, pages 173--189. Springer, 2018.

\bibitem[Choi et~al.(2018)Choi, Jang, and Alemi]{choi2018generative}
H.~Choi, E.~Jang, and A.~A. Alemi.
\newblock Waic, but why? generative ensembles for robust anomaly detection.
\newblock \emph{arXiv preprint arXiv:1810.01392}, 2018.

\bibitem[Dilokthanakul et~al.(2016)Dilokthanakul, Mediano, Garnelo, Lee,
  Salimbeni, Arulkumaran, and Shanahan]{Dilokthanakul2016DeepUC}
N.~Dilokthanakul, P.~A.~M. Mediano, M.~Garnelo, M.~J. Lee, H.~Salimbeni,
  K.~Arulkumaran, and M.~Shanahan.
\newblock Deep unsupervised clustering with gaussian mixture variational
  autoencoders.
\newblock \emph{ArXiv}, abs/1611.02648, 2016.

\bibitem[Hafner et~al.(2018)Hafner, Tran, Irpan, Lillicrap, and
  Davidson]{hafner2018reliable}
D.~Hafner, D.~Tran, A.~Irpan, T.~Lillicrap, and J.~Davidson.
\newblock Reliable uncertainty estimates in deep neural networks using noise
  contrastive priors.
\newblock \emph{arXiv preprint arXiv:1807.09289}, 2018.

\bibitem[Hendrycks et~al.(2019)Hendrycks, Mazeika, and
  Dietterich]{hendrycks2018deep}
D.~Hendrycks, M.~Mazeika, and T.~G. Dietterich.
\newblock Deep anomaly detection with outlier exposure.
\newblock \emph{International Conference on Learning Representations (ICLR)},
  2019.

\bibitem[Horé and Ziou(2010)]{Hor2010Image}
A.~Horé and D.~Ziou.
\newblock Image quality metrics: Psnr vs. ssim.
\newblock In \emph{International Conference on Pattern Recognition}, 2010.

\bibitem[Huang et~al.(2019)Huang, Dai, Nguyen, Baraniuk, and
  Anandkumar]{huang2019out}
Y.~Huang, S.~Dai, T.~Nguyen, R.~G. Baraniuk, and A.~Anandkumar.
\newblock Out-of-distribution detection using neural rendering generative
  models.
\newblock \emph{arXiv preprint arXiv:1907.04572}, 2019.

\bibitem[Kingma and Dhariwal(2018)]{kingma2018glow}
D.~P. Kingma and P.~Dhariwal.
\newblock Glow: Generative flow with invertible 1x1 convolutions.
\newblock In \emph{Advances in Neural Information Processing Systems}, pages
  10215--10224, 2018.

\bibitem[Kingma and Welling(2014)]{kingma2013auto}
D.~P. Kingma and M.~Welling.
\newblock Auto-encoding variational bayes.
\newblock \emph{International Conference on Learning Representations (ICLR)},
  2014.

\bibitem[Lakshminarayanan et~al.(2017)Lakshminarayanan, Pritzel, and
  Blundell]{lakshminarayanan2017simple}
B.~Lakshminarayanan, A.~Pritzel, and C.~Blundell.
\newblock Simple and scalable predictive uncertainty estimation using deep
  ensembles.
\newblock In \emph{Advances in neural information processing systems
  (NeurIPS)}, 2017.

\bibitem[Lee et~al.(2017)Lee, Lee, Lee, and Shin]{lee2017training}
K.~Lee, H.~Lee, K.~Lee, and J.~Shin.
\newblock Training confidence-calibrated classifiers for detecting
  out-of-distribution samples.
\newblock \emph{arXiv preprint arXiv:1711.09325}, 2017.

\bibitem[Lee et~al.(2018)Lee, Lee, Lee, and Shin]{lee2018simple}
K.~Lee, K.~Lee, H.~Lee, and J.~Shin.
\newblock A simple unified framework for detecting out-of-distribution samples
  and adversarial attacks.
\newblock In \emph{Advances in Neural Information Processing Systems
  (NeurIPS)}, 2018.

\bibitem[Liang et~al.(2018)Liang, Li, and Srikant]{liang2017enhancing}
S.~Liang, Y.~Li, and R.~Srikant.
\newblock Enhancing the reliability of out-of-distribution image detection in
  neural networks.
\newblock \emph{International Conference on Learning Representations (ICLR)},
  2018.

\bibitem[Maal{\o}e et~al.(2019)Maal{\o}e, Fraccaro, Li{\'e}vin, and
  Winther]{maaloe2019biva}
L.~Maal{\o}e, M.~Fraccaro, V.~Li{\'e}vin, and O.~Winther.
\newblock Biva: A very deep hierarchy of latent variables for generative
  modeling.
\newblock \emph{arXiv preprint arXiv:1902.02102}, 2019.

\bibitem[Malinin and Gales(2018)]{malinin2018predictive}
A.~Malinin and M.~Gales.
\newblock Predictive uncertainty estimation via prior networks.
\newblock In \emph{Advances in Neural Information Processing Systems
  (NeurIPS)}, pages 7047--7058, 2018.

\bibitem[Nalisnick et~al.(2019{\natexlab{a}})Nalisnick, Matsukawa, Teh, Gorur,
  and Lakshminarayanan]{nalisnick2018deep}
E.~Nalisnick, A.~Matsukawa, Y.~W. Teh, D.~Gorur, and B.~Lakshminarayanan.
\newblock Do deep generative models know what they don't know?
\newblock \emph{International Conference on Learning Representations (ICLR)},
  2019{\natexlab{a}}.

\bibitem[Nalisnick et~al.(2019{\natexlab{b}})Nalisnick, Matsukawa, Teh, and
  Lakshminarayanan]{nalisnick2019detecting}
E.~Nalisnick, A.~Matsukawa, Y.~W. Teh, and B.~Lakshminarayanan.
\newblock Detecting out-of-distribution inputs to deep generative models using
  a test for typicality.
\newblock \emph{arXiv preprint arXiv:1906.02994}, 2019{\natexlab{b}}.

\bibitem[Nalisnick and Smyth(2017)]{Nalisnick2017LearningAO}
E.~T. Nalisnick and P.~Smyth.
\newblock Learning approximately objective priors.
\newblock \emph{Proceedings of the 33rd Conference on Uncertainty in Artificial
  Intelligence}, 2017.

\bibitem[Ostrovski et~al.(2017)Ostrovski, Bellemare, van~den Oord, and
  Munos]{ostrovski2017count}
G.~Ostrovski, M.~G. Bellemare, A.~van~den Oord, and R.~Munos.
\newblock Count-based exploration with neural density models.
\newblock In \emph{Proceedings of the 34th International Conference on Machine
  Learning-Volume 70}, pages 2721--2730. JMLR. org, 2017.

\bibitem[Ran et~al.(2020)Ran, Xu, Mei, Xu, and Liu]{Ran2020DetectingOS}
X.~Ran, M.~Xu, L.~Mei, Q.~Xu, and Q.~Liu.
\newblock Detecting out-of-distribution samples via variational auto-encoder
  with reliable uncertainty estimation.
\newblock \emph{ArXiv}, abs/2007.08128, 2020.

\bibitem[Ren et~al.(2019)Ren, Liu, Fertig, Snoek, Poplin, Depristo, Dillon, and
  Lakshminarayanan]{ren2019likelihood}
J.~Ren, P.~J. Liu, E.~Fertig, J.~Snoek, R.~Poplin, M.~Depristo, J.~Dillon, and
  B.~Lakshminarayanan.
\newblock Likelihood ratios for out-of-distribution detection.
\newblock In \emph{Advances in Neural Information Processing Systems
  (NeurIPS)}, 2019.

\bibitem[Rezende et~al.(2014)Rezende, Mohamed, and
  Wierstra]{rezende2014stochastic}
D.~J. Rezende, S.~Mohamed, and D.~Wierstra.
\newblock Stochastic backpropagation and approximate inference in deep
  generative models.
\newblock In \emph{International Conference on Machine Learning}, pages
  1278--1286, 2014.

\bibitem[Serr{\`a} et~al.(2020)Serr{\`a}, {\'A}lvarez, G{\'o}mez, Slizovskaia,
  N{\'u}{\~n}ez, and Luque]{serra2019input}
J.~Serr{\`a}, D.~{\'A}lvarez, V.~G{\'o}mez, O.~Slizovskaia, J.~F.
  N{\'u}{\~n}ez, and J.~Luque.
\newblock Input complexity and out-of-distribution detection with
  likelihood-based generative models.
\newblock \emph{International Conference on Learning Representations (ICLR)},
  2020.

\bibitem[Song et~al.(2019)Song, Song, and Ermon]{song2019unsupervised}
J.~Song, Y.~Song, and S.~Ermon.
\newblock Unsupervised out-of-distribution detection with batch normalization.
\newblock \emph{arXiv preprint arXiv:1910.09115}, 2019.

\bibitem[Tomczak and Welling(2018)]{tomczak2017vae}
J.~Tomczak and M.~Welling.
\newblock Vae with a vampprior.
\newblock In \emph{International Conference on Artificial Intelligence and
  Statistics}, pages 1214--1223, 2018.

\bibitem[Van~den Oord et~al.(2016)Van~den Oord, Kalchbrenner, Espeholt,
  Vinyals, Graves, et~al.]{van2016conditional}
A.~Van~den Oord, N.~Kalchbrenner, L.~Espeholt, O.~Vinyals, A.~Graves, et~al.
\newblock Conditional image generation with pixelcnn decoders.
\newblock In \emph{Advances in neural information processing systems}, pages
  4790--4798, 2016.

\bibitem[Watanabe(2010)]{watanabe2010asymptotic}
S.~Watanabe.
\newblock Asymptotic equivalence of bayes cross validation and widely
  applicable information criterion in singular learning theory.
\newblock \emph{Journal of Machine Learning Research}, 11\penalty0
  (Dec):\penalty0 3571--3594, 2010.

\bibitem[Xu et~al.(2018)Xu, Chen, Zhao, Li, Bu, Li, Liu, Zhao, Pei, Feng,
  et~al.]{xu2018unsupervised}
H.~Xu, W.~Chen, N.~Zhao, Z.~Li, J.~Bu, Z.~Li, Y.~Liu, Y.~Zhao, D.~Pei, Y.~Feng,
  et~al.
\newblock Unsupervised anomaly detection via variational auto-encoder for
  seasonal kpis in web applications.
\newblock In \emph{Proceedings of the 2018 World Wide Web Conference}, pages
  187--196. International World Wide Web Conferences Steering Committee, 2018.

\end{thebibliography}

}

\newpage
\appendix
\section{Data preprocessing for different datasets}
\begin{figure}[H]
\centering
\subfigure[CIFAR10]{\includegraphics[width=0.45\textwidth]{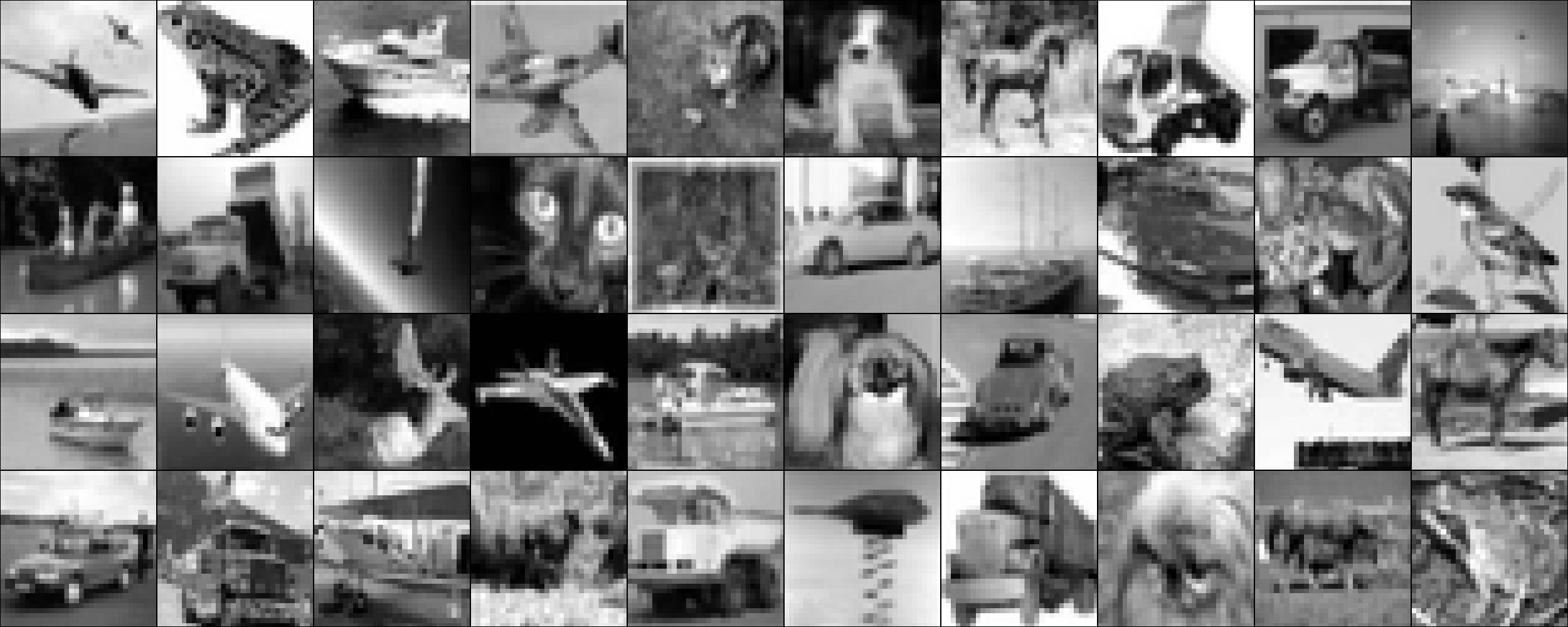}} 
\subfigure[SVHN]{\includegraphics[width=0.45\textwidth]{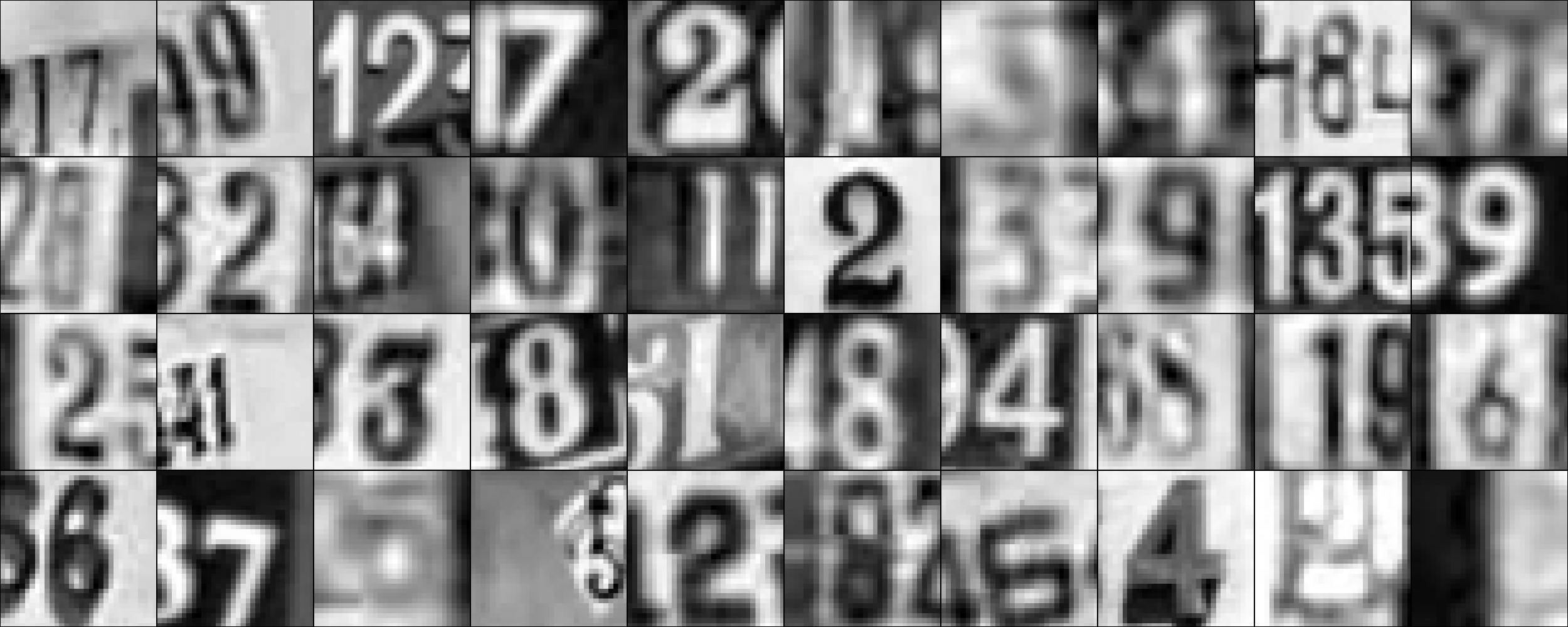}}\\
\subfigure[GTSRB]{\includegraphics[width=0.45\textwidth]{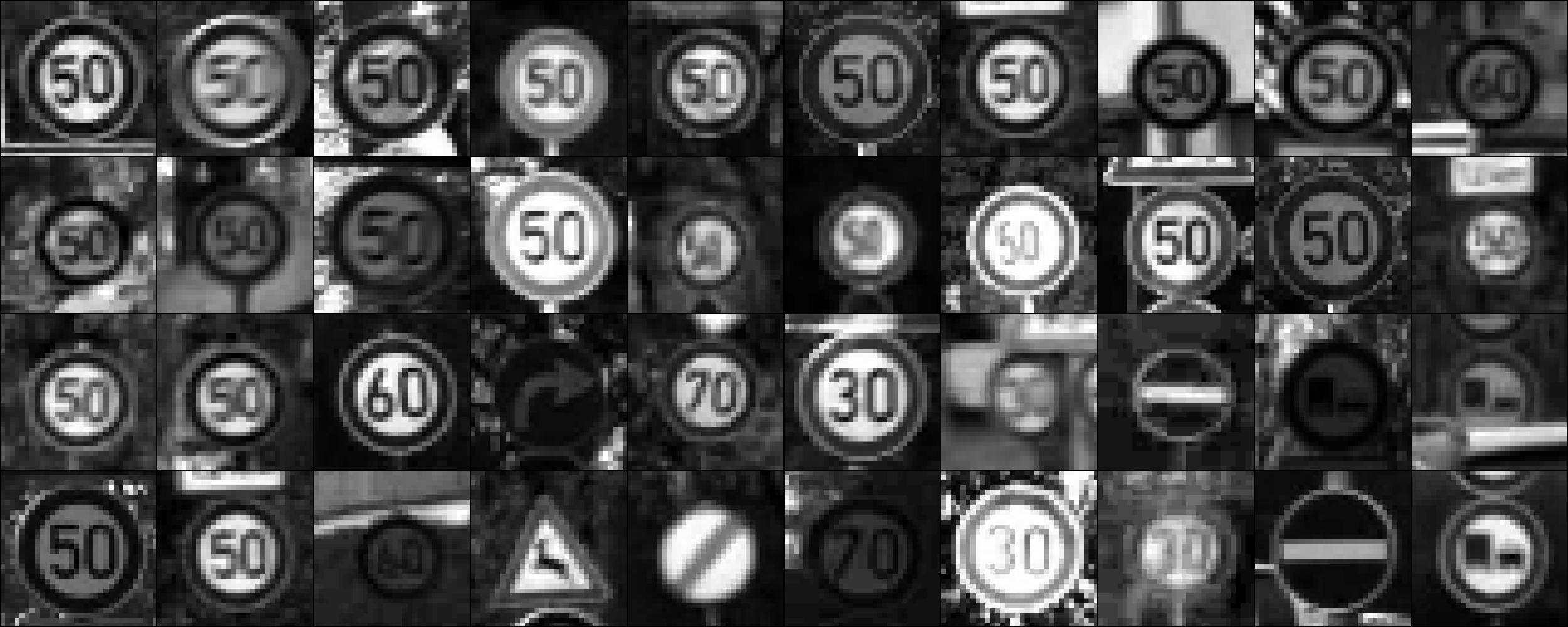}} 
\subfigure[IMAGENET]{\includegraphics[width=0.45\textwidth]{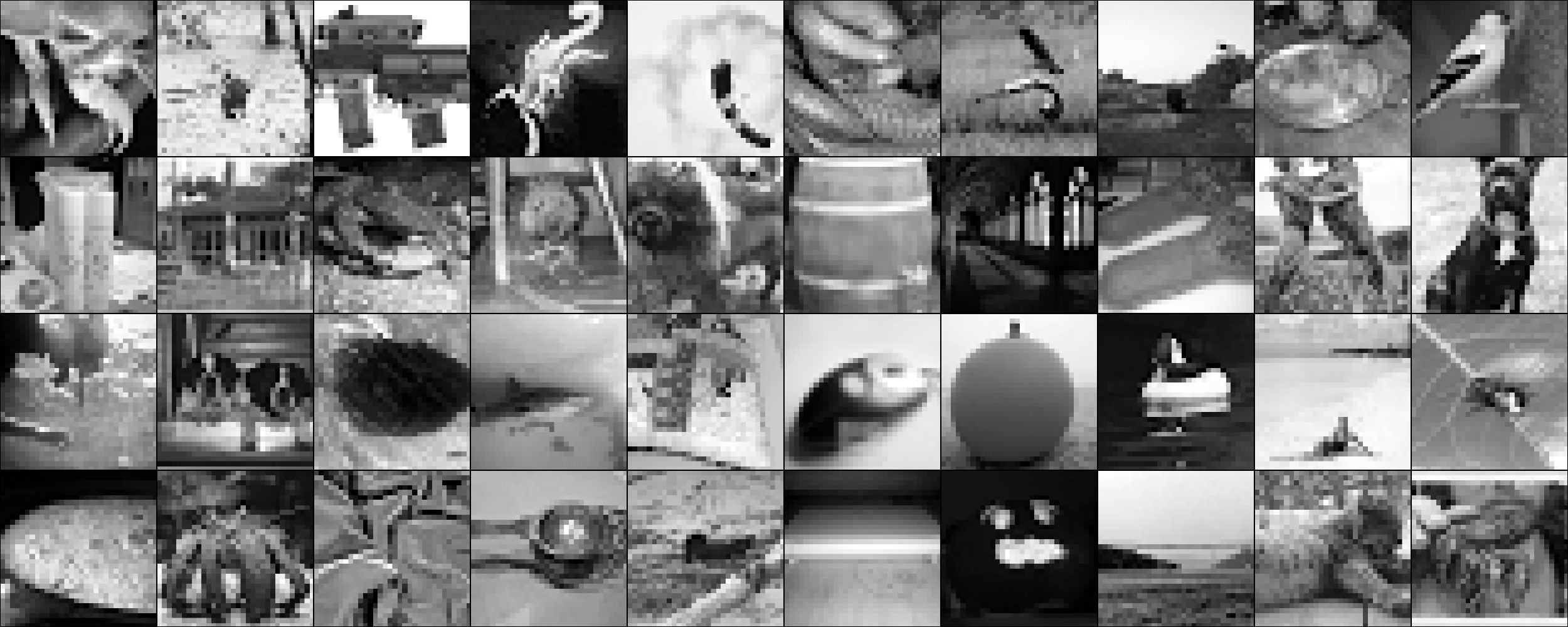}}\\
\subfigure[KMNIST]{\includegraphics[width=0.45\textwidth]{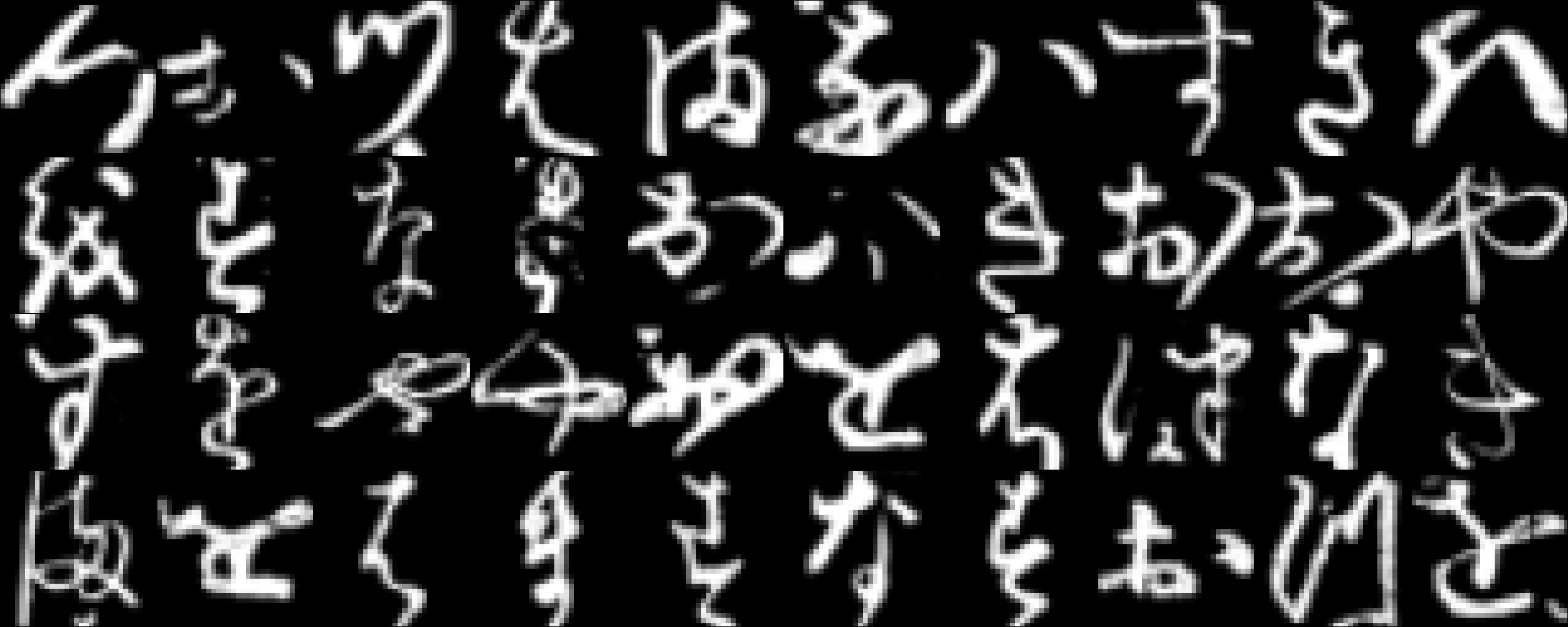}} 
\subfigure[MNIST]{\includegraphics[width=0.45\textwidth]{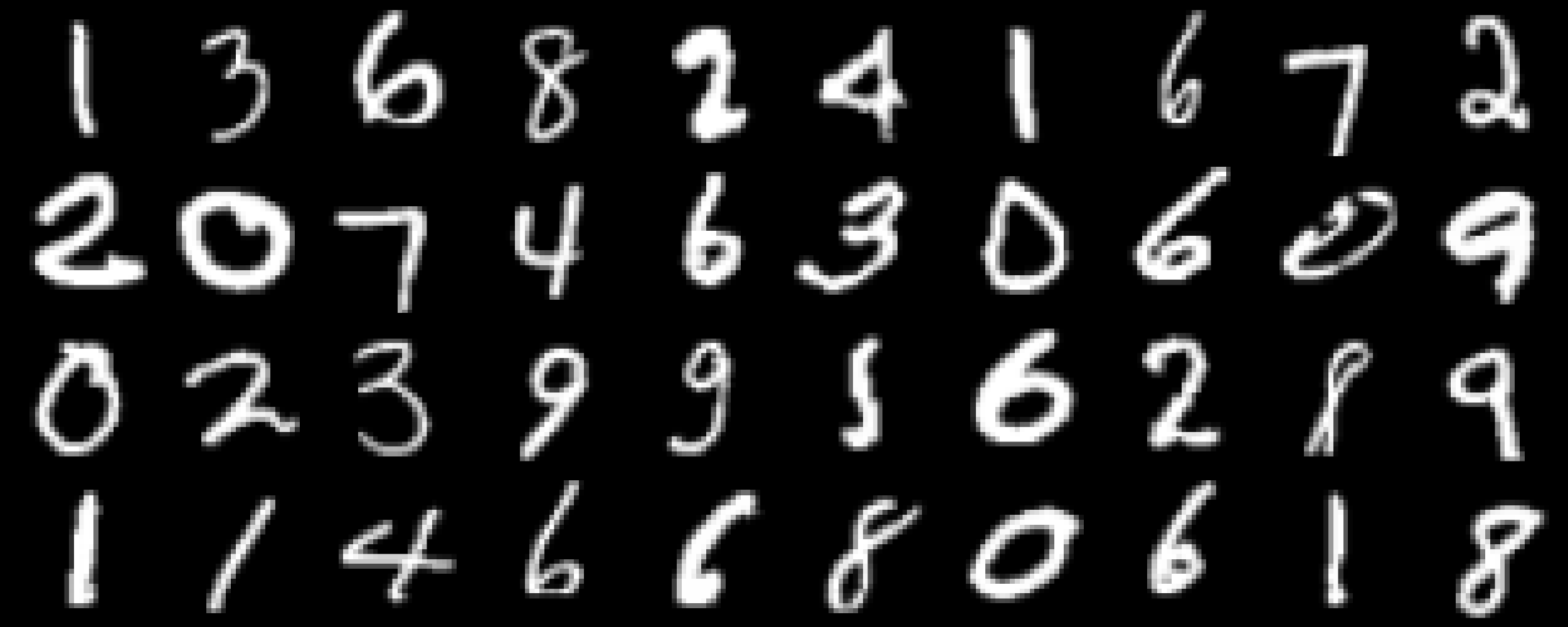}}\\
\subfigure[OMNIGLOT]{\includegraphics[width=0.45\textwidth]{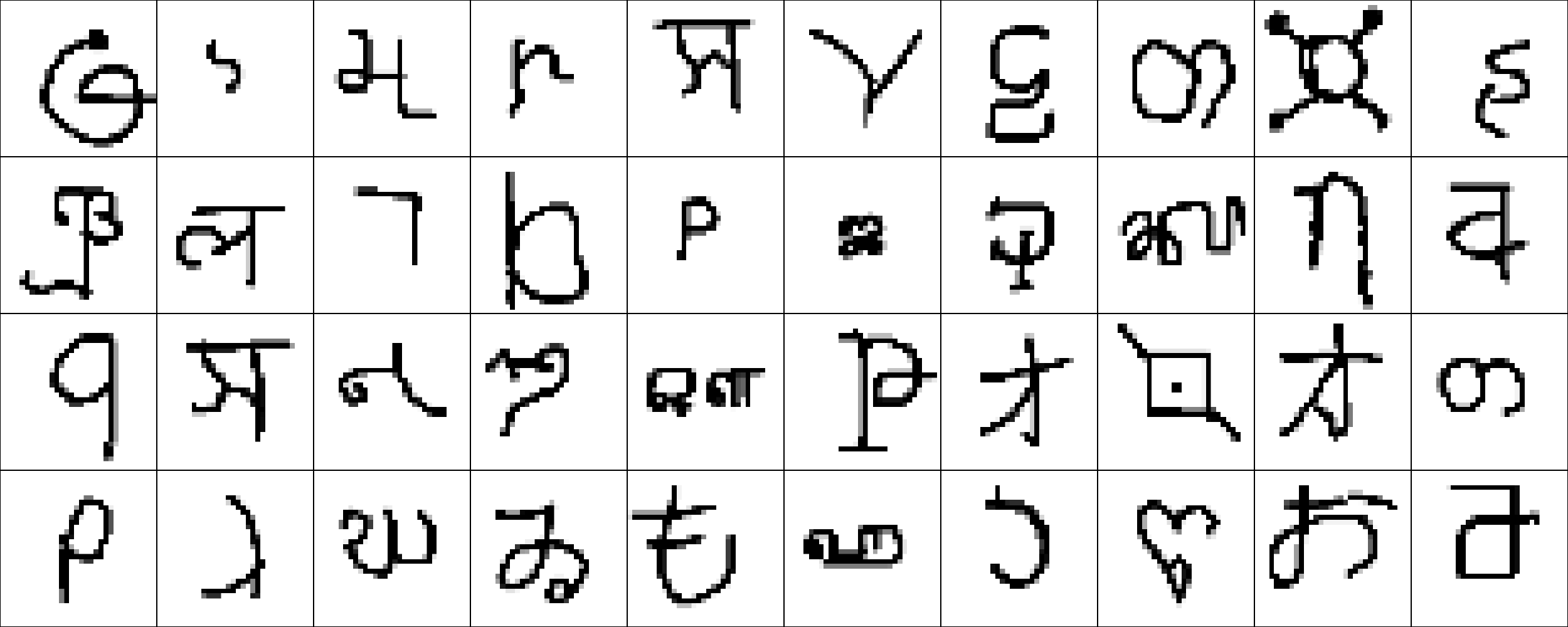}}
\subfigure[FashionMNIST]{\includegraphics[width=0.45\textwidth]{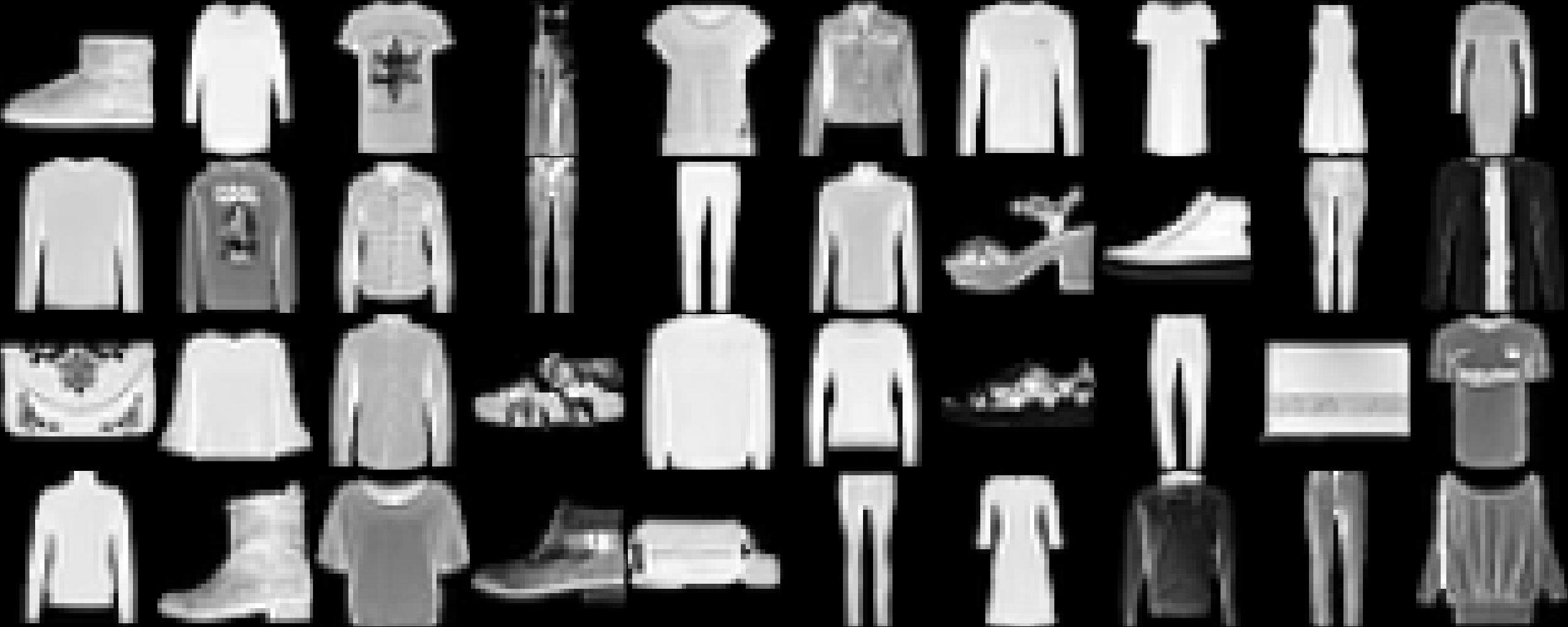}}\\
\caption[]{Presentation of preprocessed data from eight datasets, which are CIFAR10, SVHN, GTSRB, IMAGENET, KMNIST, MNIST, OMNIGLOT and FashionMNIST.}
\label{data_preprocessing}
\vspace{0.2in}
\end{figure}

We conducted some preprocessing operations in order to unify the training condition for different datasets. For datasets with RGB images, such as CIFAR10, GTSRB, IMAGENET,SVHN, we transformed these data into gray images with single channel. Then we reshaped all the images into the size (32, 32, 1). We visualized the preprocessed data from eight datasets in
\textbf{Figure~\ref{data_preprocessing}}. It is evident that the data from CIFAR10, GTSRB,IMAGENET and SVHN have the higher complexity then that from simpler datasets such as FashionMNIST, OMNIGLOT, KMNIST and MNIST.

\section{log-likelihoods of other datasets}

\begin{figure}[H]
\centering
\subfigure[Trained on GTSRB]{\includegraphics[width=0.45\textwidth]{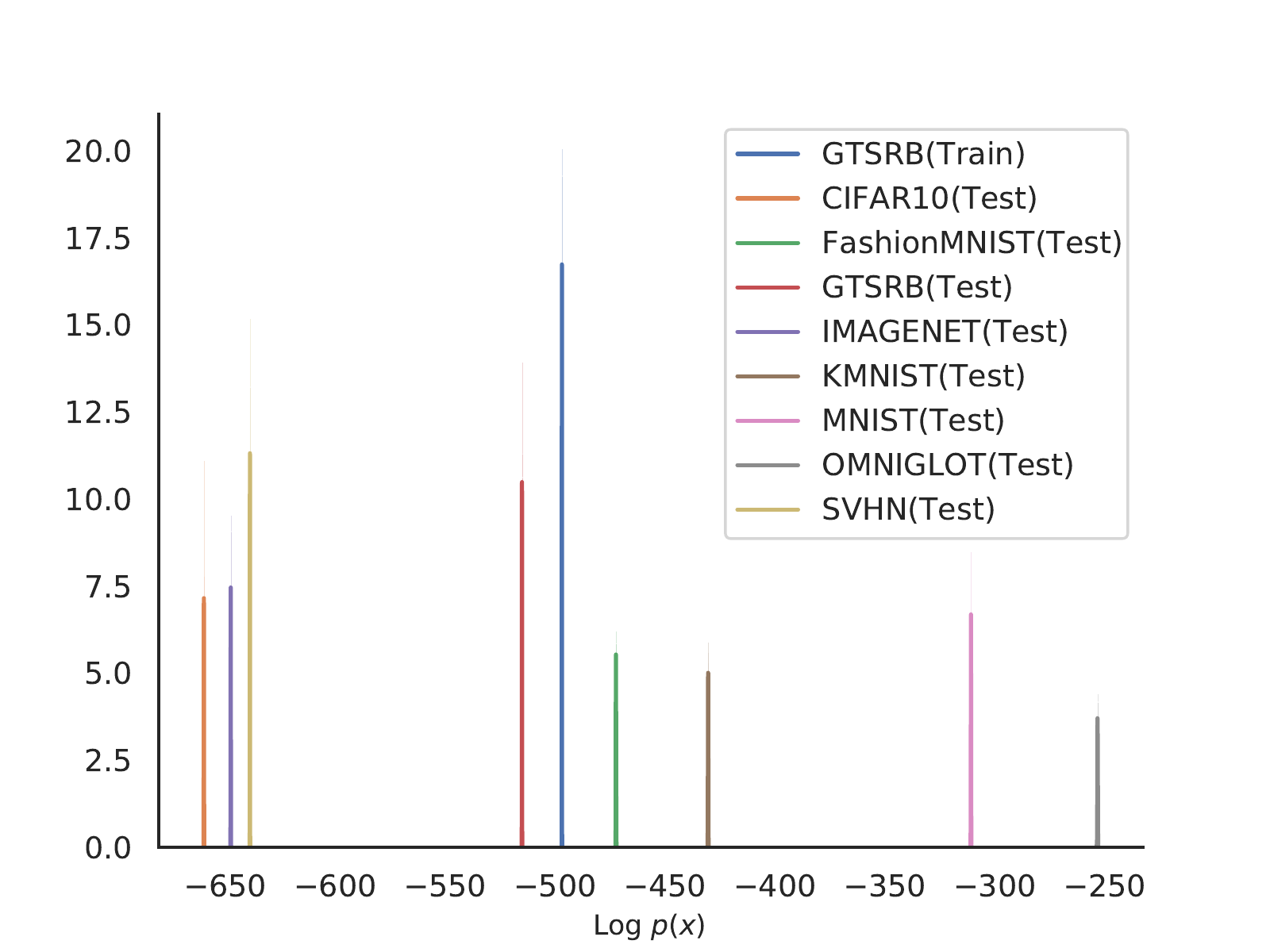}} 
\subfigure[Trained on IMAGENET]{\includegraphics[width=0.45\textwidth]{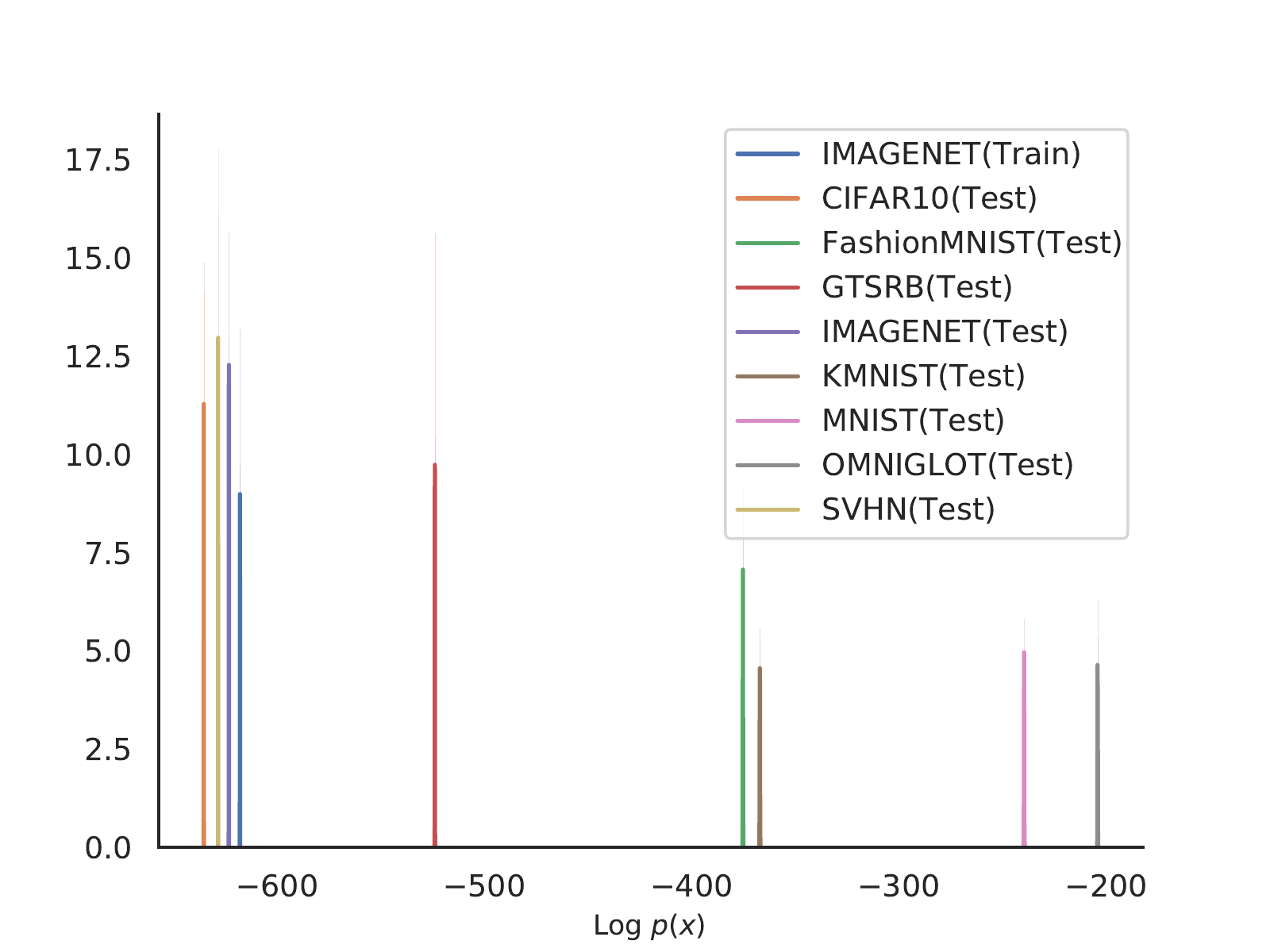}}\\
\subfigure[Trained on KMNIST]{\includegraphics[width=0.45\textwidth]{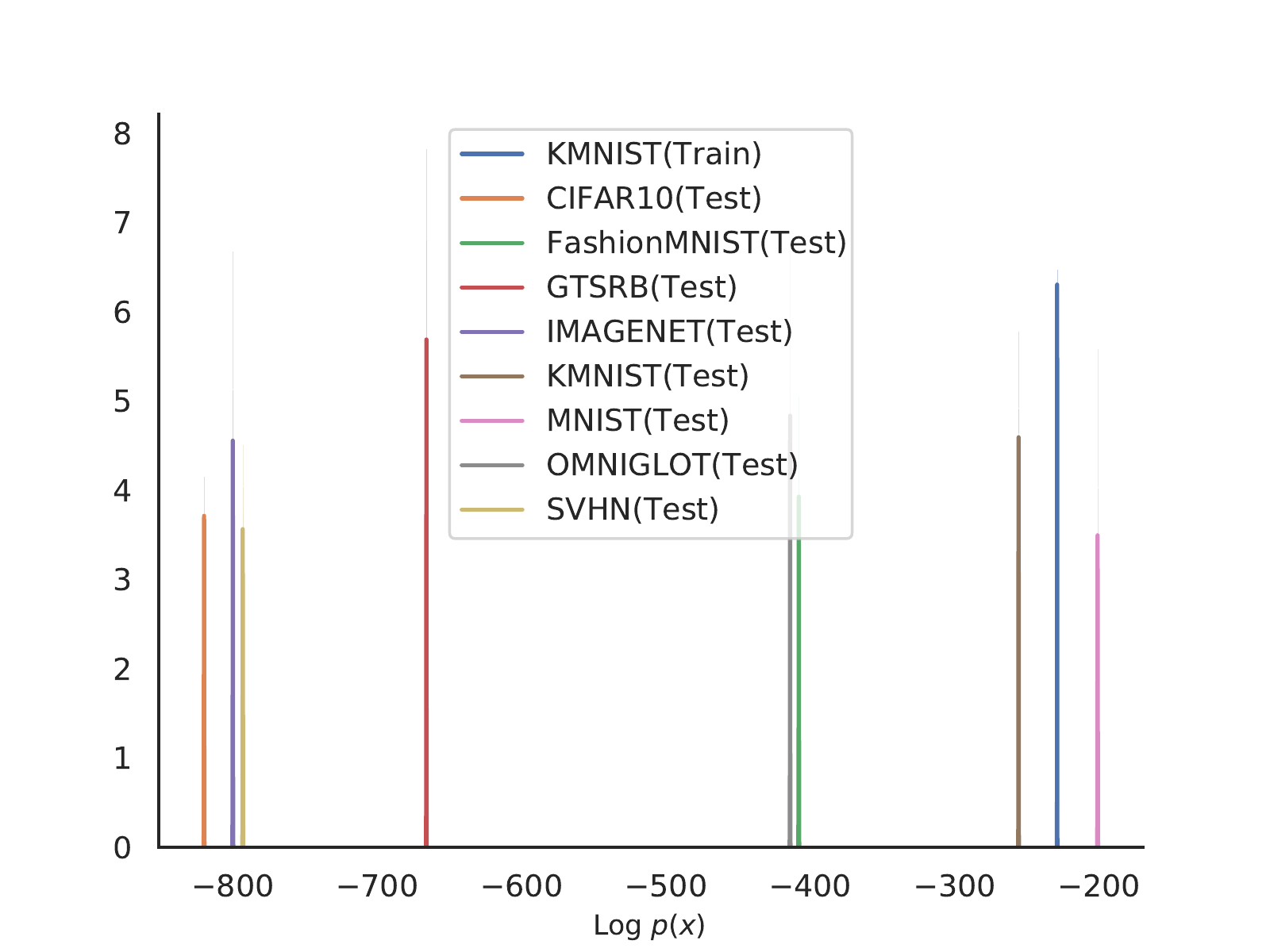}} 
\subfigure[Trained on MNIST]{\includegraphics[width=0.45\textwidth]{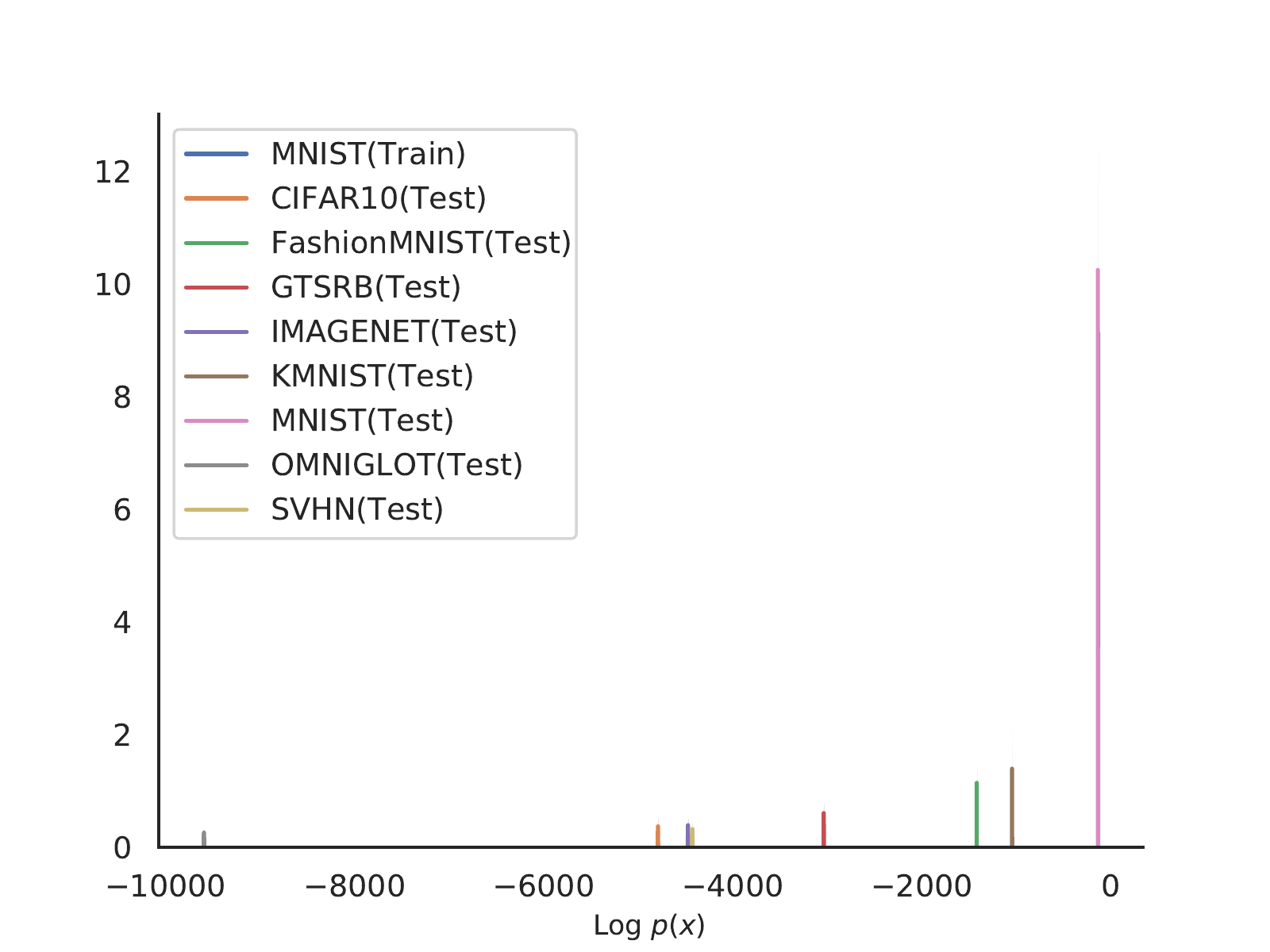}}\\

\caption[]{Histogram of log-likelihoods from a VAE model trained on GTSRB, IMAGENET, KMNIST and MNIST.}
\label{fig_trn_likelihood_appre}
\vspace{0.2in}
\end{figure}

\section{Settings for Implementation Detail}

In the experiments, VAE and BPVAE are trained with images normalized to $[0,1]$ on 1 $\times$ NVIDIA TITAN RTX GPU. In all experiments, VAE and BPVAE consist of an encoder with the architecture given in ~\cite{Ran2020DetectingOS} and a decoder shown in ~\cite{Ran2020DetectingOS}. Both VAE and BPVAE use Leaky Relu activation function. We train the VAE and BPVAE for 200 epochs with a constant learning rate $1e^{-4}$, meanwhile using Adam optimizer and batch size 64 in each experiment. 

\end{document}